\RequirePackage[OT1]{fontenc}
\documentclass[10pt, journal, compsoc]{IEEEtran}

\usepackage[utf8]{inputenc}
\usepackage{graphicx}
\usepackage{amsmath}
\usepackage{amssymb}
\usepackage{mathtools}
\usepackage{balance}
\usepackage{makecell}
\usepackage{tikz}
\usetikzlibrary{arrows.meta, backgrounds, calc, fit, shapes.misc}
\usepackage{makecell}
\usepackage{multirow}
\usepackage{booktabs}
\usepackage{pifont}
\usepackage{underscore} \usepackage{ifthen}

\usepackage[shortcuts]{extdash}
\usepackage[inline]{enumitem}
\setlist*[enumerate]{label=\emph{\roman*)}}
\usepackage{adjustbox}

\usepackage[hidelinks,
    pdftitle={Continuous Management of Machine Learning-Based Application Behavior},
    pdfauthor={Marco Anisetti, Claudio A. Ardagna, Nicola Bena, Ernesto Damiani, Paolo G. Panero}]{hyperref}

\newtheorem{definition}{Definition}\newtheorem{example}{Example}

\makeatletter\begin{document}

\title{Continuous Management of Machine Learning-Based Application Behavior}

\author{Marco Anisetti~\IEEEmembership{Senior Member,~IEEE,} Claudio A. Ardagna~\IEEEmembership{Senior Member,~IEEE,} Nicola Bena~\IEEEmembership{Student Member,~IEEE,} Ernesto Damiani~\IEEEmembership{Senior Member,~IEEE,} Paolo G. Panero\IEEEcompsocitemizethanks{\IEEEcompsocthanksitem Authors 
are with the Department of Computer Science, Universit\`a degli Studi di Milano, Milano,
Italy. E. Damiani is also with C2PS, Computer Science Department, Khalifa University, Abu Dhabi, UAE. \protect\\
E-mail: \{firstname.lastname\}@unimi.it, paologiovanni.panero@studenti.unimi.it
}
\thanks{}}

\markboth{Accepted for publication in IEEE Transactions on Services Computing; DOI: \href{https://doi.org/10.1109/TSC.2024.3486226}{10.1109/TSC.2024.3486226}}{DOI: \href{https://doi.org/10.1109/TSC.2024.3486226}{10.1109/TSC.2024.3486226}}

\IEEEcompsoctitleabstractindextext{\begin{abstract}
	Modern applications are increasingly driven by Machine Learning (ML) models whose non-deterministic behavior is affecting the entire application life cycle from design to operation. The pervasive adoption of ML is urgently calling for approaches that guarantee a stable non-functional behavior of ML-based applications over time and across model changes. To this aim, non\-/functional properties of ML models, such as privacy, confidentiality, fairness, and explainability, must be monitored, verified, and maintained. 
	Existing approaches mostly focus on \emph{i)}~implementing solutions for classifier selection according to the functional behavior of ML models, \emph{ii)}~finding new algorithmic solutions, such as continuous re\-/training. In this paper, we propose a multi-model approach that aims to guarantee a stable non-functional behavior of ML-based applications. 
	An architectural and methodological approach is provided to compare multiple ML models showing similar non-functional properties and select the model supporting stable non-functional behavior over time according to (dynamic and unpredictable) contextual changes. Our approach goes beyond the state of the art by providing a solution that continuously guarantees a stable non-functional behavior of ML-based applications, is ML algorithm-agnostic, and is driven by non-functional properties assessed on the ML models themselves. It consists of a two-step process working during application operation, where \emph{model assessment} verifies non-functional properties of ML models trained and selected at development time, and \emph{model substitution} guarantees continuous and stable support of non-functional properties. We experimentally evaluate our solution in a real-world scenario focusing on non-functional property fairness.
\end{abstract}
}

\maketitle

\IEEEpeerreviewmaketitle

\newcommand{\K}{\ensuremath{K}}
\newcommand{\va}{\ensuremath{v_s}}
\newcommand{\y}{\ensuremath{y}}
\newcommand{\partheta}{\ensuremath{\theta}}
\newcommand{\vatheta}{\ensuremath{v_s(\theta)}}

\newcommand{\twoParamFunc}[3]{\ifthenelse{\equal{#3}{}}{\ensuremath{#1_{#2}}}{\ensuremath{#1_{#2, #3}}}}
\newcommand{\paralphaSymbol}{\ensuremath{\alpha}}
\newcommand{\parbetaSymbol}{\ensuremath{\beta}}
\newcommand{\paralpha}[2]{\twoParamFunc{\paralphaSymbol}{#1}{#2}}
\newcommand{\parbeta}[2]{\twoParamFunc{\parbetaSymbol}{#1}{#2}}
\newcommand{\wat}{\ensuremath{w_{st}}}
\newcommand{\p}{\ensuremath{p}}
\newcommand{\modelsin}{\ensuremath{models}}
\newcommand{\data}{\ensuremath{traces}}
\newcommand{\ranking}{\ensuremath{models\_rank}}
\newcommand{\fnalpha}{\ensuremath{par\_alpha}}
\newcommand{\burnin}{\ensuremath{burn\_in}}
\newcommand{\winneridx}{\ensuremath{winner\_idx}}
\newcommand{\estmodels}{\ensuremath{est\_models}}
\newcommand{\draws}{\ensuremath{draws}}
\newcommand{\mc}{\ensuremath{g}}
\newcommand{\pwinner}{\ensuremath{p(\theta | y_t)}}
\newcommand{\sample}{\emph{sample}}
\newcommand{\karmbanditm}{\ensuremath{\bf DWMAB-M}}
\newcommand{\thompsonsampm}{\ensuremath{\bf thompson\_sampling}}
\newcommand{\productionm}{\ensuremath{\bf send\_into\_production}}
\newcommand{\assurancemgmntm}{\ensuremath{\bf assurance\_management}}
\newcommand{\fairpredict}{\ensuremath{\bf fair\_prediction}}
\newcommand{\scorecheck}{\ensuremath{\bf score\_check}}
\newcommand{\mcsimulation}{\ensuremath{\bf monte\_carlo\_simulation}}
\newcommand{\shouldterminate}{\ensuremath{\bf should\_terminate}}
\newcommand{\handlewindow}{\ensuremath{\bf handle\_window}}
\newcommand{\production}{\ensuremath{\bf send\_into\_production}}
\newcommand{\assurancemgmnt}{\ensuremath{\bf assurance\_management}}
\newcommand{\myspace}{\hspace{10pt}}
\newcommand{\KARMBANDIT}{\ensuremath{\bf DWMAB-M}}
\newcommand{\THOMPSONSAMP}{\ensuremath{\bf THOMPSON\_SAMPLING}}
\newcommand{\PRODUCTION}{\ensuremath{\bf SEND\_INTO\_PRODUCTION}}
\newcommand{\ASSURANCEMGMNT}{\ensuremath{\bf ASSURANCE\_MANAGEMENT}}
\newcommand{\VARIANCETHRESHOLD}{\textsc{VarianceThreshold}}
\newcommand{\FAIRPREDICT}{\ensuremath{\bf FAIR\_PREDICTION}}
\newcommand{\SCORECHECK}{\ensuremath{\bf SCORE\_CHECK}}
\newcommand{\MCSIMULATION}{\ensuremath{\bf MONTE\_CARLO\_SIMULATION}}
\newcommand{\SHOULDTERMINATE}{\ensuremath{\bf SHOULD\_TERMINATE}}
\newcommand{\HANDLEWINDOW}{\ensuremath{\bf HANDLE\_WINDOW}}
\newcommand{\variancem}{\ensuremath{var}}
\newcommand{\myitem}{\ensuremath{trace}}
\newcommand{\myarm}{\ensuremath{m^*}}
\newcommand{\anarm}{\ensuremath{model}}
\newcommand{\andm}{\ensuremath{AND}}
\newcommand{\argmax}{\ensuremath{\arg\max}}
\newcommand{\mymax}{\ensuremath{\max}}
\newcommand{\rows}{\ensuremath{rows}}
\newcommand{\mrow}{\ensuremath{row}}
\newcommand{\myalphas}{\ensuremath{alphas}}
\newcommand{\mybetas}{\ensuremath{betas}}
\newcommand{\mcdraws}{\ensuremath{100}}
\newcommand{\mccount}{\ensuremath{count}}
\newcommand{\mccounts}{\ensuremath{counts}}
\newcommand{\predictions}{\ensuremath{predicted\_vals}}
\newcommand{\valuesremain}{\ensuremath{r^{(g)}}}
\newcommand{\mpercentile}{\ensuremath{percentile}}
\newcommand{\earlysub}{\ensuremath{earlysub}}
\newcommand{\s}[1]{\ensuremath{s_{#1}}}
\newcommand{\SC}[1]{\ensuremath{S^C_{#1}}}
\newcommand{\cert}[1]{\ensuremath{C_{#1}}}
\newcommand{\certc}[1]{\ensuremath{C'_{#1}}}
\newcommand{\block}{\ensuremath{{\cal B}}}
\newcommand{\static}{\ensuremath{{\cal V}}}
\newcommand{\bool}{\ensuremath{{\cal B}}}
\newcommand{\reason}[1]{\ensuremath{R_{#1}}}
\newcommand{\phase}[1]{\ensuremath{phase_{#1}}}

\newcommand{\Cab}[1]{\ensuremath{{\mathcal C}_{#1}}}
\newcommand{\pij}[1]{\var{p}\ensuremath{_{#1}}}

\newcommand{\name}[1]{\ensuremath{\hat{#1}}}

\newcommand{\evid}[1]{\var{e}\ensuremath{_{#1}}}

\newcommand{\m}{\ensuremath{\it m}}
\newcommand{\mv}{\ensuremath{\it m}^*}

\newcommand{\score}[1]{\var{S}\ensuremath{_{#1}}}

\newcommand{\MAIN}{\ensuremath{\bf MAIN}}

\newcommand{\return}{\com{return}}
\newcommand{\push}[1]{\com{push}({#1})}
\newcommand{\myand}{\ensuremath{\wedge}}
\newcommand{\myor}{\ensuremath{\vee}}
\newcommand{\mynull}{\mbox{\sc null\/}}
\newcommand{\true}{\mbox{\sc true\/}}
\newcommand{\false}{\mbox{\sc false\/}}
\newcommand{\var}[1]{{\it #1\/}}
\newcommand{\com}[1]{\mbox{\bf #1\/}}
\newcommand{\mem}[1]{\ensuremath{\delta_{#1}}}
\newcommand{\err}{\ensuremath{\xi}}
\newcommand{\uno}{\>}
\newcommand{\due}{\>\>}
\newcommand{\tre}{\>\>\>}
\newcommand{\quattro}{\>\>\>\>}
\newcommand{\5}{\>\>\>\>\>}
\newcommand{\6}{\>\>\>\>\>\>}

\newcommand{\commentall}[1]{\emph{/* #1*/}}
\newcommand{\commentopen}[1]{\emph{/* #1}}
\newcommand{\commentclose}[1]{\emph{#1*/}}
\newcommand{\commentin}[1]{\emph{#1}}

\newcommand{\INPUT}{{\bf INPUT}}
\newcommand{\OUTPUT}{{\bf OUTPUT}}

\newcommand{\property}{\ensuremath{p}}
\newcommand{\executionTrace}{\ensuremath{et}}
\newcommand{\MLModel}{\ensuremath{m}}
\newcommand{\service}{\ensuremath{s}}
\newcommand{\selected}[1]{\ensuremath{\hat{#1}}}
\newcommand{\optimum}[1]{#1\ensuremath{^\ast}}
\newcommand{\optimumTime}[2]{#1\ensuremath{^{\ast,#2}}}

\newcommand{\candidateList}{\ensuremath{cl}}

\newcommand{\assuranceLevelSymbol}{\ensuremath{al}}
\newcommand{\assuranceLevel}[1]{\ensuremath{\assuranceLevelSymbol_{#1}}}

\newcommand{\ok}{\ding{51}}
\newcommand{\notok}{\ding{55}}
\newcommand{\soSo}{\ensuremath{\approx}}
\newcommand{\NA}{--}

\newcommand{\objectiveStability}{Behavior stab.}
\newcommand{\objSelection}{Select the best ensemble for a (set of) data point wrt quality metrics}

\newcommand{\window}{\ensuremath{w}}

\newcommand{\thresholdAssurance}{\ensuremath{thr}}
\newcommand{\degradationSymbol}{deg}
\newcommand{\degradation}[1]{\ensuremath{\text{\degradationSymbol}_{#1}}}

\newcommand{\dynamicMABVariableWindow}{DMVW}
\newcommand{\dynamicMABVariableWindowMemory}{DMVW-Mem}

\newcommand{\categorySelection}{S}
\newcommand{\categoryAdaptationFunctional}{FA}
\newcommand{\categoryAdaptationNonFunctional}{NFA}

\newcommand{\relatedWorkHeaderObjTypeSubFunc}{Func.}
\newcommand{\relatedWorkHeaderObjTypeSubNonFunc}{Non-Func.}

\newcommand{\relatedWorkHeaderGenericitySubModel}{ML Model}
\newcommand{\relatedWorkHeaderGenericitySubProperty}{Property}

\newcommand{\relatedWorkHeaderCategoryShort}{Cat.}
\newcommand{\relatedWorkHeaderCategoryLong}{Category}
\newcommand{\relatedWorkHeaderObjShort}{Objective}
\newcommand{\relatedWorkHeaderObjTypeShort}{Objective Type}
\newcommand{\relatedWorkHeaderGenericityShort}{Applicability}
\newcommand{\relatedWorkHeaderObjLong}{Objective}
\newcommand{\relatedWorkHeaderObjTypeLong}{Objective Type}
\newcommand{\relatedWorkHeaderGenericityLong}{Applicability}

\newcommand{\regret}{\ensuremath{r}}
\newcommand{\rankingMetricSymbol}{\ensuremath{rm}}
\newcommand{\rankingMetric}[2]{\twoParamFunc{\rankingMetricSymbol}{#1}{#2}}

\newcommand{\instant}{\ensuremath{t}}

\newcommand{\reward}{\ensuremath{v}}
\newcommand{\rewdist}[1]{\ensuremath{f_\MLModel{#1}(\y{} \mid \theta)}}

\newcommand{\thresholdRegret}{\ensuremath{\text{\emph{residual}}_\regret}}

\newcommand{\expSymbol}{\emph{exp}}
\newcommand{\experiment}[1]{\ensuremath{\text{\expSymbol}_{#1}}}

\newcommand{\penaltyFunc}{\ensuremath{\text{\emph{penalty}}}}
\newcommand{\rankingFunc}{\ensuremath{R}}

\newcommand{\BetaDistribution}[2]{\ensuremath{\twoParamFunc{\text{Beta}}{#1}{#2}}}

\newcommand{\objectiveTypeFunctional}{F}
\newcommand{\objectiveTypeNonFunctional}{NF}

\newcommand{\funcName}[1]{\textbf{#1}}

\newcommand{\cumulative}{\ensuremath{cum\_assurance}}
\newcommand{\cumulativeError}{\ensuremath{\hat{\err}}} 
\section{Introduction}\label{sec:intro}
Machine Learning (ML) has become the technique of choice to provide advanced functionalities and carry out tasks hardly achievable by traditional control and optimization algorithms~\cite{DA.SOFSEM2020}. Even the behavior, orchestration, and deployment parameters of distributed systems and services, possibly offered on the cloud-edge continuum, are increasingly based on ML models~\cite{10.1145/3341145}. 
Concerns about the black-box nature of ML have led to a societal push that involves all components of society (policymakers, regulators, academic and industrial stakeholders, citizens) towards trustworthy and transparent ML, giving rise to legislative initiatives on artificial intelligence (e.g., the AI Act in Europe~\cite{eu-aia}). 

This scenario introduces the need for solutions that continuously guarantee a stable non-functional behavior of ML-based applications, a task that is significantly more complex than mere QoS\-/based selection and composition (e.g.,~\cite{https://doi.org/10.1111/exsy.13602, 9187421,Yoshioka2019}). The focus of such a task is to assess the non-functional properties of ML models, such as privacy, confidentiality, fairness, and explainability, over time and across changes. The non-functional assessment of ML\-/based applications behavior has to cope with the ML models' complexity, low transparency, and continuous evolution~\cite{AABD.IC2023, BreckerLinsSunyaev20231000166100}. ML models in fact are affected by model and data drifts, quality degradation, and accuracy loss, which may substantially impact on the quality and soundness of the application itself.

Recent research points to solutions where ML models evolve according to contextual changes (e.g., a shift in the incoming data distribution), typically via continuous re\-/training and peculiar training algorithms and ML models~\cite{pmlr-v139-tahmasbi21a, 10.5555/3367032.3367242, 9288346}. Other solutions consider \emph{classifier selection} where a (set of) ML model is statically or dynamically selected according to some criteria~\cite{ROY2018179, zhang4608310sv, Zhu2023, ALMEIDA201867}; in this context, dynamic selection identifies the most suitable ML model for each data point at inference time. Ensembles have been also considered to increase ML robustness~\cite{pmlr-v162-chen22k, jia2021intrinsic, levine2021deep, wang2022improved, AABBDY.TSUSC2023}. 
Finally, some solutions have initially discussed certification\-/based assessment of ML-based applications~\cite{AABD.IC2023, BreckerLinsSunyaev20231000166100, BAGA.SNCS2024}. Current approaches however fall short in supporting the requirements of modern ML\-/based applications. On the one hand, they disregard stable application behavior and non\-/functional properties, which are increasingly mandated by law, in favor of accuracy maximization. On the other hand, they do not provide a general solution that applies to \emph{any} non\-/functional properties and ML algorithms, and rather focus on specific, though relevant, properties (e.g., fairness) and algorithms (e.g., decision trees).

This paper fills in the above gaps by defining a multi\-/model approach that guarantees a stable non-functional behavior of ML-based applications. Similarly to dynamic classifier selection, our approach keeps a pool of ML models and one ML model at time is dynamically selected during inference according to a (set of) non\-/functional property; the selected ML model is replaced only when its non\-/functional property degrades. 
Our approach is particularly suited for constrained and critical scenarios with (dynamic and unpredictable) contextual changes. In such scenarios online re\-/training and dynamic classifier/ensemble selection approaches \begin{enumerate*}
\item have a larger overhead due to the expensive training and the need to select a model \emph{for each data point}, and
\item can lead to unexpected application behavior due to the arrival of new, unpredictable input data. 
\end{enumerate*} 

Our multi\-/model approach is built on a two-step process working during application operation as follows. The first step, \emph{model assessment}, verifies non-functional properties of ML models already trained and selected at development time. To this aim, we extend our previous work on Multi\-/Armed Bandit (MAB)~\cite{AADP.MEDES2020}, towards a dynamic MAB that assesses non\-/functional properties of ML models at run time. The second step, \emph{model substitution}, is driven by the properties assessed at step \emph{i)}, and guarantees a stable support for non-functional properties over time and across changes. Our approach can be used both as a complete solution for application behavior management (from design to operation), or to complement existing ML-based applications with a multi-model substitution approach.

Our contribution is threefold.
We first propose a new definition of non\-/functional property of ML models. Our definition departs from traditional, attribute\-/based properties available in the literature (e.g.,~\cite{AAB.TSC2023, AB.SSE2023}), and includes a scoring function at the basis of ML models comparison and selection. We extend
the scope of traditional properties, which are mostly based on accuracy~\cite{ROY2018179} or metrics unrelated to the model itself (e.g., the battery level of the device or latency~\cite{276950}), to include non\-/functional properties such as fairness and integrity, often mandated by law. Though important, these properties are often neglected in literature~\cite{doshivelez2017rigorous}.
We then describe our multi-model approach for managing the non-functional behavior of ML-based applications. Our approach defines a dynamic MAB for the assessment of the non-functional properties of ML models, and proposes two model substitution strategies built on it. The two strategies support the dynamic choice of the model with the best set of non\-/functional properties at run time, by ranking and substituting the models in a dynamically sized evaluation window, and performing additional early substitutions upon severe non\-/functional degradation using an assurance-based evaluation.
We note that, although a plethora of assurance techniques exist for the verification of non\-/functional properties in traditional service\-/based applications~\cite{AAB.TSC2023, ABBJ.ICWS2022, AADV.CSUR2015, 7393762}, the definition of rigorous assurance\-/based processes for ML-based applications is still more an art than a science~\cite{AABD.IC2023, BreckerLinsSunyaev20231000166100}. 
We finally extensively evaluate our solution focusing on non\-/functional property \emph{fairness}.

The remainder of this paper is organized as follows. Section~\ref{sec:scenario} presents our reference scenario and our approach at a glance. Section~\ref{sec:buildingblocks} describes our building blocks, including Static MAB that is later extended in Section~\ref{sec:DMAB} towards Dynamic MAB for non-functional ML model assessment. Section~\ref{sec:substitution} presents the two strategies for model substitution. Section~\ref{sec:walkthrough} describes our approach in an end\-/to\-/end walkthrough. Section~\ref{sec:experiments} presents an extensive experimental evaluation in a real scenario. Section~\ref{sec:related} comparatively discusses the related work and Section~\ref{sec:conclusion} draws our conclusions.
 \section{Our Approach at a Glance}\label{sec:scenario}

We consider a scenario where a service provider is willing to deploy an application (service workflow) whose behavior depends on an ML model. The service provider needs to maintain stable performance across time in terms of quality (e.g., accuracy of the model) and non-functional posture (e.g., fairness).
Let us assume a scenario where the model behavior changes, such as model drift (e.g., due to online partial or full re-training) or data drift (e.g., service re-deployment or migration in the cloud\-/edge continuum), which are induced by modifications in the application operational conditions. To cope with this scenario, the service provider adopts a multi-model approach by designing and deploying multiple models that can be alternatively used depending on the context. This multi-model deployment can impact single or multiple nodes in cloud or cloud-edge scenarios. We note that the model behavior is evaluated at design time and continuously monitored at run time to decide which model to use during application operation.
We also note that the service provider can decide to substitute or not the model in operation due to restrictions in the application environment, but is always capable of comparing the behavior of the model in use with the other alternative models and use this evidence to fine-tune them offline.

\begin{figure}[!t]
	\begin{adjustbox}{max totalsize={.95\columnwidth}{\textheight},center}
		\begin{tikzpicture}
    \tikzset{
        target/.style={
            circle,
            draw,
            minimum size=1cm,
            inner sep=0pt,
            thin,
            fill=white
        },
        target service/.style={
            chamfered rectangle,
            chamfered rectangle angle=30,
            draw,
            thin,
            inner sep=2.5pt,
            minimum width=1.3cm,
            minimum height=1.3cm
        },
        cloud target/.style={
            double distance={.2cm}
        },
        process step/.style={
            rectangle,
            draw,
            rounded corners,
            thin
        },
        every node/.append style={
            font=\scriptsize
        },
        link/.style={
            >=Latex,
        },
        link node basic/.style={
            font=\scriptsize,
            inner sep=1pt,
        },
        on link/.style={
            link node basic,
            fill=white,
        },
        container/.style={
            very thin,
            dashed,
            draw
        },
        caption layer/.style={
            font=\normalsize,
            inner sep=0pt
        },
        pics/service with ml/.style n args={3}{
            code={
\node[target service, fill=white] (target-#1) at (#2) {};
                \node[target] (target-#1-inner) at (target-#1){\makecell{#3}};
            }
        },
        mab/.style={
            rectangle,
            draw
        }
    }

    \newcommand{\labelUpperShift}{10}
    \newcommand{\foldingShift}{3.5}

    \begin{scope}[on background layer]
         \node[target] (target-1-1) {};
    \end{scope}

    \begin{scope}[on background layer]
        \node[target] at ([xshift=\foldingShift, yshift=-\foldingShift]target-1-1) (target-1-2) {};
   \end{scope}

    \node[target] at ([xshift=\foldingShift, yshift=-\foldingShift]target-1-2) (target-1-3) {\makecell{Model\\$\MLModel_1$}};

    \pic{service with ml={2-1}{[xshift=75]target-1-3.east}{$\selected{\MLModel}$}};

\pic{service with ml={3-1}{[xshift=70]target-2-1.east}{$\hat{\MLModel}'$}};
    \begin{scope}[on background layer]
        \node[target service, fill=none] at ([xshift=\foldingShift, yshift=\foldingShift]target-3-1) {};
    \end{scope}

\node[process step] (step-1) at ([yshift=-80]target-1-3.south) {\makecell{Model\\Substitution}};

    \node[process step] (step-2) at (step-1 -| target-2-1) {\makecell{Multi-Model\\Approach}};

    \node[process step] (step-3) at (step-2 -| target-3-1) {\makecell{Dynamic MAB}};

\coordinate (midway between property and services) at ($(step-3)!.5!(target-3-1-inner)$);
   \node[process step, yshift=-7.5] (mab) at (midway between property and services) {\makecell{Model Assesment}};

   \draw[->, link] (mab) to node[on link]{based on} (step-3);

\draw[->, link] (step-1) to node[on link] {selects} (target-1-3);
    \draw[->, link] (step-2) to node[on link] {triggers} (step-1);
    \draw[->, link, bend right=17.5] (step-2.east) to node[on link] {uses} (mab.south west);
    \draw[->, link] (target-1-3) to node[on link] {substitutes} (target-2-1);
    \draw[->, link] (step-2.north west) to node[on link, yshift=10] {\makecell{ranks candidates\\models}} (target-1-3); 
\begin{scope}[on background layer]
        \draw[->, link] (step-2) to node[on link, yshift=-10] {\makecell{evaluates Assurance\\level \assuranceLevelSymbol}} (target-2-1);
   \end{scope}
   \draw[<->, link, double] (target-2-1) to (target-3-1);

\node[] (target 1 caption)  at ([yshift=\labelUpperShift]target-1-1.north) {\makecell{Ranked Candidate\\List \candidateList}};
   \node[] (target 2 caption) at (target 1 caption -| target-2-1) {\makecell{ML\-/Based\\Service}};
   \node[] (target 3 caption) at (target 2 caption -| target-3-1) {\makecell{Application\\Services}};

\coordinate (midway between target 1 caption and target 2 caption coordinate) at ([yshift=25]$(target 1 caption)!.5!(target 2 caption)$);
   \node[caption layer] (caption edge) at (midway between target 1 caption and target 2 caption coordinate) {\makecell{\textbf{Edge or Cloud Node$_\text{1}$}}}; 

   \coordinate (midway between caption edge and target 3 caption coordinate) at (caption edge -| target 3 caption);
   \node[caption layer] (caption cloud) at (midway between caption edge and target 3 caption coordinate) {\makecell{\textbf{Edge or Cloud}\\\textbf{Node$_\text{n}$}}};

    \begin{scope}[on background layer]
        \node[container, fit=(target-1-1)(target-2-1)(target 1 caption)(target 2 caption)(caption edge)] (container-1) {};
\end{scope}

    \begin{scope}[on background layer]
        \node[container, fit=(target-3-1)(target 3 caption)(caption cloud)] (container-2) {};
\end{scope}

    \coordinate (midway between target2 and target3) at ($(target-2-1)!.5!(target-3-1)$);
\coordinate (midway between cloud 1 and cloud 2) at ($(container-1.east)!.5!(container-2.west)$); \path let \p1 = (midway between cloud 1 and cloud 2), \p2 = ($(mab.north)!.5!(target-3-1.south)$) in
        coordinate (execution trace capture position) at (\x1, \y2);
    \node[on link] (execution trace capture) at (execution trace capture position) {\makecell{captures execution\\traces \executionTrace}};
 
\path let \p1 = (midway between target2 and target3), \p2 = (midway between cloud 1 and cloud 2) in
        coordinate (line between midway between cloud and execution trace capture position) at (\x2, \y1);
    \draw[-, link] (line between midway between cloud and execution trace capture position) to (execution trace capture);

    \draw[->, link, bend left=22.5] (execution trace capture) to (step-2.north east);

\path let \p1 = (midway between cloud 1 and cloud 2), \p2 = ($(midway between cloud 1 and cloud 2)!.5!(container-1.north)$) in
    coordinate (dots position) at (\x1, \y2);

    \node[] (dots) at (dots position) {\vdots};

\end{tikzpicture}

 	\end{adjustbox}
	\caption{Overview of our approach. \label{fig:overview}} 
\end{figure}
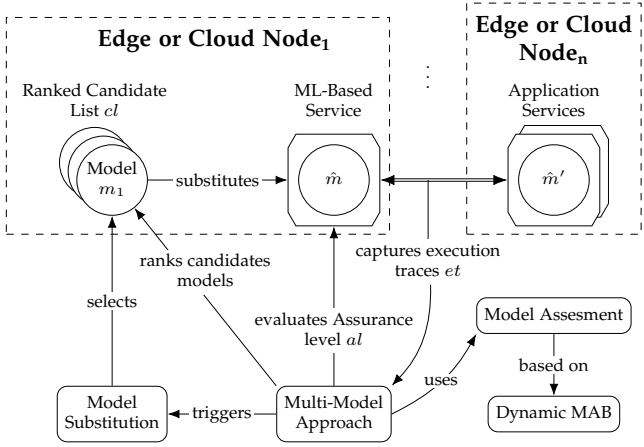

Figure~\ref{fig:overview} shows an overview of the above scenario and how we apply our multi\-/model approach to address the continuous management of ML-based application non-functional behavior.

Our approach starts at development time with a set of pre-trained, candidate ML models and statically selects the model with the best (set of) non-functional property to be used by the application. At run time, two processes, namely, \emph{model assessment} and \emph{model substitution}, continuously monitor the non\-/functional property(ies) of all models and apply model substitution when necessary to maintain stable application behavior. The two processes work in an \emph{evaluation window}.

Let \candidateList\ denote the set of candidate models $\{\MLModel_0,$ $\ldots,$ $\MLModel_k$$\}$ and \selected{\MLModel} the model currently in use. 
Process \emph{model assessment} (Section~\ref{sec:DMAB}) evaluates models in \candidateList\ according to the given non\-/functional property \property. It implements a Dynamic Multi-Armed Bandit (\emph{Dynamic MAB}) approach, which extends our previous work built on traditional MAB~\cite{AADP.MEDES2020} to continuously evaluate the models.

Process \emph{model substitution} (Section~\ref{sec:substitution}) takes as input the results of process model assessment and selects the best model \selected{\MLModel} to be used within the application according to two strategies.
The first strategy compares models in \candidateList\ using the Dynamic MAB in the entire evaluation window, producing a model ranking. The best model in the ranking is then selected as the new \selected{\MLModel} to be used by the application in the following evaluation window. 
The second strategy extends the first one by implementing \emph{early substitutions} of \selected{\MLModel} according to metric assurance level \assuranceLevelSymbol, measuring the model degradation. Early substitutions anticipate the replacement of \selected{\MLModel}, addressing transient changes before the end of the evaluation window. 

\begin{example}[Reference Scenario]\label{ex:scenario}
	Our reference scenario considers an ML\-/based application that supports authorities (i.e., courts) in estimating the bail of an individual in prison. The application trains 5 models \candidateList$=$$\{\MLModel_1,$ \ldots$,$ $\MLModel_5\}$ in the cloud on the same dataset, containing data on past bails at national level. Each court is then provided with a model. Let us assume that the selected model is $\MLModel_3$ (i.e., \selected{\MLModel}$=$$\MLModel_3$). Due to the nature of the task, the non\-/functional property of interest is \emph{fairness}, in terms of variance over some protected attributes~\cite{AADP.MEDES2020}.

Let us assume that, at run time, \selected{\MLModel} shows significant biases 
in the presence of underrepresented/disadvantaged groups, thus affecting predicted bails. 
	The overall fairness of \selected{\MLModel} must be evaluated and compared to the other candidate models, and a model substitution triggered when needed to maintain stable non-functional behavior.
\end{example}

Our reference scenario exemplifies the four main challenges of modern ML applications: \begin{enumerate*}
	\item the definition of advanced non\-/functional properties that are typical of ML such as fairness and privacy;
	\item the assessment and comparison of models in terms of a given non\-/functional property;
	\item the detection of an application's non\-/functional property degradation at run time;
	\item the automatic substitution of models to keep the application behavior stable with respect to their non\-/functional properties.
\end{enumerate*}
Existing solutions in literature cannot tackle these challenges in their entirety. For instance, QoS\-/aware service selection approaches (e.g.,~\cite{https://doi.org/10.1111/exsy.13602, 9187421, Hosseinzadeh2020}) maximize specific (non\-/)functional metrics to build an optimum composition or retrieve the most suitable models. Similarly, classifier selection approaches (e.g.,~\cite{ROY2018179, zhang4608310sv, Zhu2023, ALMEIDA201867}) maximize quality metrics such as accuracy, continuously swapping the models and potentially introducing fluctuations in the non\-/functional behavior. Other approaches target non\-/AI systems (e.g.,~\cite{AADDEP.TSC2021}), or do not generalize over the ML algorithms or non\-/functional properties (e.g.,~\cite{10.5555/3367032.3367242, 9288346, 10.1007/978-3-030-27615-720, 10.1007/978-3-030-75765-620, iosifidis2021online}).
To the best of our knowledge, our multi\-/model approach is the first solution that guarantees \emph{stable application non\-/functional behavior over time} and is \emph{generic with respect to the ML algorithm and property}. A detailed comparison of the approach in this paper with solutions in literature is provided in Section~\ref{sec:related}.
 \section{Building Blocks}\label{sec:buildingblocks}
Our multi\-/model approach is based on three main building blocks: \begin{enumerate*}
    \item \emph{execution traces} (Section~\ref{subsec:buildingblocks-traces}),
    \item \emph{non\-/functional properties} (Section~\ref{subsec:buildingblocks-property}), and
    \item \emph{Multi-Armed Bandit} (Section~\ref{subsec:buildingblocks-staticmab}). 
\end{enumerate*}
Table~\ref{tbl:symbols} shows the terminology used in this paper.

\begin{table}[!t]
    \centering
    \footnotesize
    \caption{Terminology. \label{tbl:symbols}}
    \begin{adjustbox}{max totalsize={.99\columnwidth}{\textheight},center}
		\begin{tabular}{lp{.8\columnwidth}}
    \toprule
    \textbf{Symbol} & \textbf{Description}\\
    \midrule
    \MLModel & ML model\\
    \candidateList & Set of candidate ML models\\
    \selected{\MLModel} & Selected ML model\\
    \executionTrace$_\instant$ & \instant\-/th execution trace \\
    \pij{} & Non-functional property\\
    \pij{}.\name{\pij{}} & Non-functional property name\\
    \pij{}.\score{} & Non-functional property score function\\
    \optimum{\MLModel}{} & ML model providing the highest reward according to the MAB \\
    \window & Variable\-/sized sequence of observed execution traces (window) \\
    $\BetaDistribution{\MLModel}{j}$ & Beta distribution of model \MLModel\ in window $\window_j$\\
    \paralpha{\MLModel}{j} and \parbeta{\MLModel}{j} & Parameters of \BetaDistribution{\MLModel}{j}\\
    $\reward_\MLModel(\theta)$ & Reward retrieved from \BetaDistribution{\MLModel}{j} \\ 
    \thresholdRegret & Threshold of residual value in the MAB for closing the window \window\\
\mem{} & Memory to re\-/initialize $\BetaDistribution{\MLModel}{j}$ in a new window \window\\
    $\rankingMetric{\MLModel}{\instant}$ & Value of ranking metric \rankingMetricSymbol\ of \MLModel\ at the \instant\-/th execution trace \\
    $\assuranceLevel{\instant}$ & Assurance level of \selected{\MLModel} at the \instant\-/th execution trace\\ $\degradation{\instant}$ & Degradation of the assurance level of \selected{\MLModel} up to the \instant\-/th execution trace \\
    \thresholdAssurance & Threshold of assurance level degradation triggering early substitution \\
    \bottomrule
\end{tabular} 	\end{adjustbox}
\end{table}

\subsection{Execution Traces}\label{subsec:buildingblocks-traces}
Execution traces capture the behavior of a given ML model at run time. They can be defined as follows.

\begin{definition}[Execution Trace]\label{def:execution-trace}
An execution trace $\executionTrace$ is a tuple of the form $\langle$\emph{dp}$,$ \emph{pred}$\rangle$ where \begin{enumerate*}
    \item \emph{dp} is the data point (i.e., a set of features) given as input to a model,
    \item \emph{pred} is the predicted result. \end{enumerate*}
\end{definition}

\begin{figure*}[!t]
	\begin{adjustbox}{max totalsize={\textwidth}{\textheight},center}
\includegraphics[width=\textwidth]{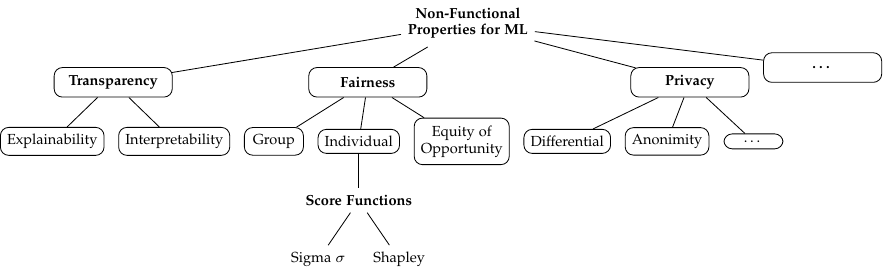}
	\end{adjustbox}
	\caption{Partial view of the ML property taxonomy~\cite{AADP.MEDES2020}.\label{fig:tax}}
\end{figure*}

We note that \emph{dp} can also contain the raw samples given as input to a deep learning model. Execution traces can be captured, for instance, by intercepting calls to the ML\-/based application or through monitoring~\cite{AAB.ICSOC2023}.

\begin{example}[Execution Trace]\label{ex:trace}
Following Example~\ref{ex:scenario}, let us consider an execution trace $\executionTrace$$=$$\langle[$\emph{age}$=$27$,$ \emph{gender}$=$male$,$ \emph{race}$=$latino$,$ \emph{past\-/offence}$=$0$,$ \ldots$],$ \texttt{\$10K}$\rangle$, retrieved by monitoring model \selected{\MLModel}, where: $[$\emph{age}$=$27$,$ \ldots$,]$ is the data point sent to \selected{\MLModel} and \texttt{\$10K} is the predicted bail.
\end{example}

\subsection{Non-Functional Properties}\label{subsec:buildingblocks-property}
Traditional non\-/functional properties are defined as an abstract property (i.e., the property name) refined by a set of attributes~\cite{AAB.TSC2023}. Common properties include \emph{performance}, \emph{confidentiality}, \emph{integrity}, \emph{availability}~\cite{softwarearchitecture}. When an ML model is considered, the notion of property is redesigned~\cite{AABD.IC2023} as follows. 

\begin{definition}[Non-Functional Property]\label{def:property}
A non\-/functional property $\pij{}$ is a pair \pij{}$=$($\name{\pij{}}$,$\score{}$), where $\hat{\pij{}}$ is an abstract property taken from a shared controlled vocabulary~\cite{AADP.MEDES2020} and \score{} is a score function of the form \score{}$\colon$$\{\executionTrace\}$$\rightarrow$$\mathbb{R}$ quantitatively describing \emph{how much} an ML model supports $\hat{\pij{}}$ according to its execution traces.
\end{definition}

In the following, we use the dotted notation to refer to the components of \pij{} (e.g., $\pij{}.\score{}$).

\begin{example}[Non-Functional Property]\label{ex:pr}
    Following Example~\ref{ex:trace}, property \emph{fairness} can be defined as \pij{\text{fairness}}$=$$($\emph{fairness}$,$ \emph{variance\-/over\-/gender\-/race}$)$, where the score function \begin{enumerate*}
        \item generates a number of synthetic data points \emph{dp} covering all the possible combinations of protected attributes \emph{gender} and \emph{race};
        \item sends each \emph{dp} to the model; and
        \item measures the variance $\sigma^2$ over the predicted bails.
    \end{enumerate*}
    We note that the higher the variance, the lower the support for property fairness.
\end{example}

Non\-/functional properties of ML can be peculiar properties purposefully defined for ML evaluation (e.g., adversarial robustness~\cite{BAGA.SNCS2024}) or a new interpretation of traditional ones (e.g., model prediction integrity)~\cite{AABD.IC2023}. Figure~\ref{fig:tax} shows a portion of our taxonomy of non\-/functional properties, which has been fully presented in our previous work~\cite{AADP.MEDES2020}. The taxonomy includes generic properties, which are then refined by detailed properties. 
For example, \emph{transparency} is a generic property with two sub-properties: \begin{enumerate*}
    \item \emph{explainability}, 
    the capability to explain the model, on one hand, and individual decisions taken by the model, on the other hand; and
    \item \emph{interpretability},  
    the capability to predict the consequences on a model when changes are observed. 
\end{enumerate*}
As another example, \emph{fairness} is a generic property with multiple sub\-/properties. For each detailed property, different score functions can be defined. For instance, Figure~\ref{fig:tax} shows two score functions for property \emph{individual fairness}: variance $\sigma$ (used in this paper) and \emph{Shapley}~\cite{10.1145/3457607}. Score functions in the taxonomy are general, though we note that they need to be refined and instantiated in the context of an evaluation process for a specific ML\-/based application. 

\subsection{The MAB} \label{subsec:buildingblocks-staticmab}

We use the Multi-Armed Bandit (MAB) technique~\cite{AADP.MEDES2020} to compare models according to a non\-/functional property \pij{} on a set of execution traces. MAB repeatedly executes an experiment, whose goal is to get the highest reward that can be earned by executing a specific action chosen among a set of alternatives. Every action returns a reward or a penalty with different (and unknown) probabilities. The experiment is commonly associated with the problem of a gambler facing different slot machines (or a single slot machine with many arms retrieving different results). In our scenario, the actions are the models $\MLModel_i$\ in the candidate list \candidateList\ and the reward is based on the score function $\pij{}.\score{}$ in Definition~\ref{def:property}.

\begin{definition}[MAB]\label{def:MAB}
    Let \candidateList\ be the set of candidate models $\{$$\MLModel_1,$ \ldots$,$ $\MLModel_k$$\}$, each associated with an unknown reward $\reward_{\MLModel}$ for non\-/functional property $\pij{}$. The goal of the MAB is to select the model \optimum{\MLModel} providing the highest reward in a set of experiments (i.e., a set of execution traces). A probability distribution \rewdist{} drives experiments' rewards, with \y{} the observed reward and \partheta{} a collection of unknown parameters that must be learned through experimentation.
MAB is based on Bayesian inference considering that, in each experiment, the success/failure odd of each model is unknown and can be shaped with the probability distribution \textit{Beta distribution}. Let \MLModel\ be a model, its Beta distribution $\text{Beta}_\MLModel$ is based on two parameters \paralphaSymbol, \parbetaSymbol$\in$$[0,$ $1]$ (denoted as \paralpha{\MLModel}{} and \parbeta{\MLModel}{}, resp.) and its probability density function can be represented as

    \begin{equation} 
        \text{Beta}_\MLModel(x; \paralpha{\MLModel}{}, \parbeta{\MLModel}{}) = \frac{x^{(\paralpha{\MLModel}{} - 1 )} (1 - x)^{\parbeta{\MLModel}{} - 1 }}{B(\paralpha{\MLModel}{}, \parbeta{\MLModel}{})}  \label{eq:betadist}
    \end{equation}
where the normalization function \ensuremath{B} is the Euler beta function

    \begin{equation} 
        B(\paralpha{\MLModel}{}, \parbeta{\MLModel}{}) = \int^{1}_{0} x^{\paralpha{\MLModel}{} - 1} (1 - x)^{\parbeta{\MLModel}{} - 1} dx  \label{eq:beta}
    \end{equation}

    Thompson sampling~\cite{Chapelle2011} pulls models in \candidateList, as a new trace \executionTrace\ is received from the application, by sampling the models' Beta distribution. The model with the highest sampled reward (denoted as \optimum{\MLModel}) is then evaluated according to $\pij{}.\score{}$ and \executionTrace. A comparison of the score function output against a threshold determines the success or failure of this evaluation. $\text{Beta}_{\optimum{\MLModel}}$ is then updated accordingly, such that \optimum{\MLModel} is \emph{pulled} more frequently in case of successful evaluation ($\paralpha{\optimum{\MLModel}}{}$ increased by $1$), less frequently ($\parbeta{\optimum{\MLModel}}{}$ increased by $1$), otherwise. 
    
    Let $\y_\instant$ denote the set of observations recorded up to the \instant\-/th execution trace $\executionTrace_t$. The optimal model \optimum{\MLModel} is selected according to probability $\text{\emph{winner}}_{\MLModel, \instant}$:

    \begin{equation}
        \begin{split}
        \text{\emph{winner}}_{\MLModel, \instant} = P(\optimum{\MLModel}\mid \y_\instant) = \\
        = \int l(\optimum{\MLModel} = \arg \max_{\MLModel\in\candidateList} \reward_m(\theta))p(\theta \mid y_\instant) d\instant \label{eq:winner}
        \end{split}
    \end{equation}

    where $l$ is the indicator function and $p(\theta$$\mid$$\y_\instant)$ is the Bayesian posterior probability distribution of $\theta$ given the observations up to the \instant\-/th execution trace.
The MAB terminates when all experiments end, that is, all traces have been received. 

\end{definition}

The optimal model \optimum{\MLModel} is used by the application (i.e., \selected{\MLModel}$=$\optimum{\MLModel})~\cite{AADP.MEDES2020}.  We note that, while effective at application startup, the MAB cannot be continuously applied at run time as new traces come.
For this reason, the MAB in this section (\emph{Static MAB} in the following) is only used for static model selection at development time. We then define in Section~\ref{sec:DMAB} a \emph{Dynamic MAB} as the extension of the Static MAB for run\-/time model selection and substitution.
 \section{Model Assessment: Dynamic MAB}\label{sec:DMAB}

Process model assessment compares ML models at run time according to their non-functional behavior. It takes as input the models in the candidate list \candidateList\ and the non-functional property \property, and returns as output the models' Beta distributions. Models assessment uses the Static MAB within an \emph{evaluation window} \window\ of $\vert \window \vert$ execution traces, and then shifts the window of $\vert \window \vert$ execution traces instantiating a new Static MAB.

The window size $\vert \window \vert$ can be fixed or variable. When $\window$ has fixed size $\vert \window \vert$, the Dynamic MAB may not reach a statistical relevance to take a decision;
 in this case, \emph{i)} the outcome can be sub-optimal or \emph{ii)} the evaluation can be extended to the next window.

 When $\window$ has variable size $\vert \window \vert$, our default approach, the MAB terminates the evaluation and moves to the next window only when a statistically relevant decision can be made. It is based on the \emph{value remaining in the experiment}~\cite{Scott2015}, a tunable strategy that controls both the estimation error and the window size requested to reach a valuable decision. In the following, we present our solutions based on variable window sizes, namely Dynamic MAB with Variable Window (\dynamicMABVariableWindow) and \dynamicMABVariableWindow\ with Memory (\dynamicMABVariableWindowMemory).

\subsection{Dynamic MAB with Variable Window (\dynamicMABVariableWindow)}\label{subsec:DMAB-W}
The Dynamic MAB with Variable Window (\dynamicMABVariableWindow) implements the \emph{value remaining in the experiment} using a Monte Carlo simulation. The simulation considers a random set \ensuremath{g} of sampled draws from models' Beta distributions. It then counts the frequency of each model being the winner in \ensuremath{g} as an estimation of the corresponding probability distribution.

The \emph{value remaining in the experiment} is based on the minimization of the ``regret'' (the missed reward) due to an early terminated experiment. Let \ensuremath{\theta_0} denote the value of \ensuremath{\theta} and \optimum{\MLModel}$=$$\arg\max_{\MLModel\in\candidateList} \reward_\MLModel(\theta_0)$ the optimal model at the end of a window $w$. The regret due to early termination of an experiment within window \window\ is represented by $\reward_{\optimum{\MLModel}}(\theta_0)$$-$$\reward_{\optimumTime{\MLModel}{\instant}}(\theta_0)$, which is the difference between \emph{i)} the reward $\reward_{\optimum{\MLModel}}(\theta_0)$ of the optimal model \optimum{\MLModel} retrieved at the end of window \window\ and \emph{ii)} the reward $\reward_{\optimumTime{\MLModel}{\instant}}(\theta_0)$ of the optimal model \optimumTime{\MLModel}{\instant} retrieved at execution trace $\executionTrace_\instant$.

Considering that the regret is not directly observable, it can be computed using the posterior probability distribution.
Let us consider $\reward_{\ast}(\theta^{(g)})$$=$$\max_{\MLModel\in\candidateList} \reward_{\MLModel}(\theta^{(g)})$ where $\theta^{(g)}$ is drawn from $p(\theta$$\mid$$ y_t)$. The ``regret'' \regret\ in $g$ is $\regret^{(g)}$$=$$\reward_{\ast}(\theta^{(g)})$$-$$\reward_{\optimumTime{\MLModel}{\instant}}(\theta_0)$, which derives from the regret posterior probability distribution. We note that $\reward_{\ast}(\theta^{(g)})$ is the maximum available value within each Monte Carlo draw set \ensuremath{g} and $\reward_{\optimumTime{\MLModel}{\instant}}(\theta^{(g)})$ is the value (alike taken in \ensuremath{g}) for the best arm within each Monte Carlo simulation. Regret is expressed as the percentage of the deviation from the model identified as the winner, so that draws from the posterior probability are given as follows. 

\begin{equation}
	\regret^{(g)} = \frac{\reward_{\ast}(\theta^{(g)})-\reward_{\optimumTime{\MLModel}{\instant}}(\theta^{(g)})}{\reward_{\optimumTime{\MLModel}{\instant}}(\theta^{(g)})}\label{eq:regret}
\end{equation}

The experiment completes when 95\% of the samples of a simulation have a residual value less than a given percentage (\thresholdRegret) of the value of the best model $\reward_{\optimumTime{\MLModel}{\instant}}(\theta_{0})$. Formally, a window can be closed when $\text{\emph{percentile}}(\regret^{(g)}, 95)$$\leq$$\reward_{\optimumTime{\MLModel}{\instant}}(\theta_0)$$\times$$\thresholdRegret$. A common value for \thresholdRegret\ is $1\%$; it can be increased to reduce the window size, while leading to a greater residual. We note that the window size can be tuned in terms of the acceptable regret using \thresholdRegret.

In a nutshell, \dynamicMABVariableWindow\ takes a decision based on the execution traces in a specific window \window\ only. A new MAB is executed from scratch in each window, potentially leading to a discontinuous model comparison. Due to this effect, \dynamicMABVariableWindow\ can produce fluctuations in the selection of the optimal model \optimum{\MLModel} to be used by the application.
To address these issues, we extend the \dynamicMABVariableWindow\ with the notion of \emph{memory} in Section~\ref{subsec:DMAB-WM}. 

\subsection{\dynamicMABVariableWindow\ with Memory (\dynamicMABVariableWindowMemory)}\label{subsec:DMAB-WM}

The \emph{\dynamicMABVariableWindow\ with memory }(\dynamicMABVariableWindowMemory) keeps track of past \dynamicMABVariableWindow\ executions to smooth the discontinuity among consecutive windows.  \dynamicMABVariableWindowMemory\ for window $\window_{j}$ is defined on the basis of the Beta distributions and corresponding parameters in window $\window_{j-1}$ as follows.

\begin{definition}[\dynamicMABVariableWindow\ with memory (\dynamicMABVariableWindowMemory)]\label{def:mem}
	A \dynamicMABVariableWindowMemory\ is a \dynamicMABVariableWindow\ where the Beta distribution $\BetaDistribution{\MLModel}{j}$ of each model \MLModel\ in window $\window_{j}$ is initialized on the basis of the Beta distribution $\BetaDistribution{\MLModel}{j-1}$ of the corresponding model \MLModel\ in window $\window_{j-1}$, as follows:

	\begin{itemize}
		\item $\paralpha{\MLModel}{j}$$=$$\paralpha{\MLModel}{j-1}\times\mem{}$
		\item $\parbeta{\MLModel}{j}$$=$$\parbeta{\MLModel}{j-1}\times\mem{}$,
	\end{itemize}

	where \mem{}$\in$$[0,$ $1]$ denotes the memory size, $\paralpha{\MLModel}{j-1}$ and $\parbeta{\MLModel}{j-1}$ are \paralphaSymbol\ and \parbetaSymbol\ of Beta distribution $\text{Beta}_{\MLModel, j-1}$ of model \MLModel\ in window $\window_{j-1}$. We note that the resulting $\paralpha{\MLModel}{j}$ and $\parbeta{\MLModel}{j}$ are rounded down and set to $1$ when equal to $0$.
\end{definition}

\dynamicMABVariableWindowMemory\ initializes the Beta distributions in each window $\window_{j}$ according to the Beta distribution parameters observed in window $\window_{j-1}$.

\begin{figure}[t]
	\begin{adjustbox}{max totalsize={.95\columnwidth}{\textheight},center}
		\begingroup
  \makeatletter
  \providecommand\color[2][]{\GenericError{(gnuplot) \space\space\space\@spaces}{Package color not loaded in conjunction with
      terminal option `colourtext'}{See the gnuplot documentation for explanation.}{Either use 'blacktext' in gnuplot or load the package
      color.sty in LaTeX.}\renewcommand\color[2][]{}}\providecommand\includegraphics[2][]{\GenericError{(gnuplot) \space\space\space\@spaces}{Package graphicx or graphics not loaded}{See the gnuplot documentation for explanation.}{The gnuplot epslatex terminal needs graphicx.sty or graphics.sty.}\renewcommand\includegraphics[2][]{}}\providecommand\rotatebox[2]{#2}\@ifundefined{ifGPcolor}{\newif\ifGPcolor
    \GPcolorfalse
  }{}\@ifundefined{ifGPblacktext}{\newif\ifGPblacktext
    \GPblacktexttrue
  }{}\let\gplgaddtomacro\g@addto@macro
\gdef\gplbacktext{}\gdef\gplfronttext{}\makeatother
  \ifGPblacktext
\def\colorrgb#1{}\def\colorgray#1{}\else
\ifGPcolor
      \def\colorrgb#1{\color[rgb]{#1}}\def\colorgray#1{\color[gray]{#1}}\expandafter\def\csname LTw\endcsname{\color{white}}\expandafter\def\csname LTb\endcsname{\color{black}}\expandafter\def\csname LTa\endcsname{\color{black}}\expandafter\def\csname LT0\endcsname{\color[rgb]{1,0,0}}\expandafter\def\csname LT1\endcsname{\color[rgb]{0,1,0}}\expandafter\def\csname LT2\endcsname{\color[rgb]{0,0,1}}\expandafter\def\csname LT3\endcsname{\color[rgb]{1,0,1}}\expandafter\def\csname LT4\endcsname{\color[rgb]{0,1,1}}\expandafter\def\csname LT5\endcsname{\color[rgb]{1,1,0}}\expandafter\def\csname LT6\endcsname{\color[rgb]{0,0,0}}\expandafter\def\csname LT7\endcsname{\color[rgb]{1,0.3,0}}\expandafter\def\csname LT8\endcsname{\color[rgb]{0.5,0.5,0.5}}\else
\def\colorrgb#1{\color{black}}\def\colorgray#1{\color[gray]{#1}}\expandafter\def\csname LTw\endcsname{\color{white}}\expandafter\def\csname LTb\endcsname{\color{black}}\expandafter\def\csname LTa\endcsname{\color{black}}\expandafter\def\csname LT0\endcsname{\color{black}}\expandafter\def\csname LT1\endcsname{\color{black}}\expandafter\def\csname LT2\endcsname{\color{black}}\expandafter\def\csname LT3\endcsname{\color{black}}\expandafter\def\csname LT4\endcsname{\color{black}}\expandafter\def\csname LT5\endcsname{\color{black}}\expandafter\def\csname LT6\endcsname{\color{black}}\expandafter\def\csname LT7\endcsname{\color{black}}\expandafter\def\csname LT8\endcsname{\color{black}}\fi
  \fi
    \setlength{\unitlength}{0.0500bp}\ifx\gptboxheight\undefined \newlength{\gptboxheight}\newlength{\gptboxwidth}\newsavebox{\gptboxtext}\fi \setlength{\fboxrule}{0.5pt}\setlength{\fboxsep}{1pt}\definecolor{tbcol}{rgb}{1,1,1}\begin{picture}(7200.00,5040.00)\gplgaddtomacro\gplbacktext{\csname LTb\endcsname \put(946,704){\makebox(0,0)[r]{\strut{}0.0}}\put(946,1161){\makebox(0,0)[r]{\strut{}5.0}}\put(946,1618){\makebox(0,0)[r]{\strut{}10.0}}\put(946,2076){\makebox(0,0)[r]{\strut{}15.0}}\put(946,2533){\makebox(0,0)[r]{\strut{}20.0}}\put(946,2990){\makebox(0,0)[r]{\strut{}25.0}}\put(946,3447){\makebox(0,0)[r]{\strut{}30.0}}\put(946,3905){\makebox(0,0)[r]{\strut{}35.0}}\put(946,4362){\makebox(0,0)[r]{\strut{}40.0}}\put(946,4819){\makebox(0,0)[r]{\strut{}45.0}}\put(1078,484){\makebox(0,0){\strut{}0.7}}\put(2032,484){\makebox(0,0){\strut{}0.8}}\put(2986,484){\makebox(0,0){\strut{}0.8}}\put(3941,484){\makebox(0,0){\strut{}0.9}}\put(4895,484){\makebox(0,0){\strut{}0.9}}\put(5849,484){\makebox(0,0){\strut{}1.0}}\put(6803,484){\makebox(0,0){\strut{}1.0}}}\gplgaddtomacro\gplfronttext{\csname LTb\endcsname \put(1210,4646){\makebox(0,0)[l]{\strut{}$\BetaDistribution{\MLModel_5}{11}(110, 2)$}}\put(1210,4426){\makebox(0,0)[l]{\strut{}$\BetaDistribution{\MLModel_5}{12}(11, 1)$}}\put(209,2761){\rotatebox{-270.00}{\makebox(0,0){\strut{}PDF}}}\put(3940,154){\makebox(0,0){\strut{}x}}}\gplbacktext
    \put(0,0){\includegraphics[width={360.00bp},height={252.00bp}]{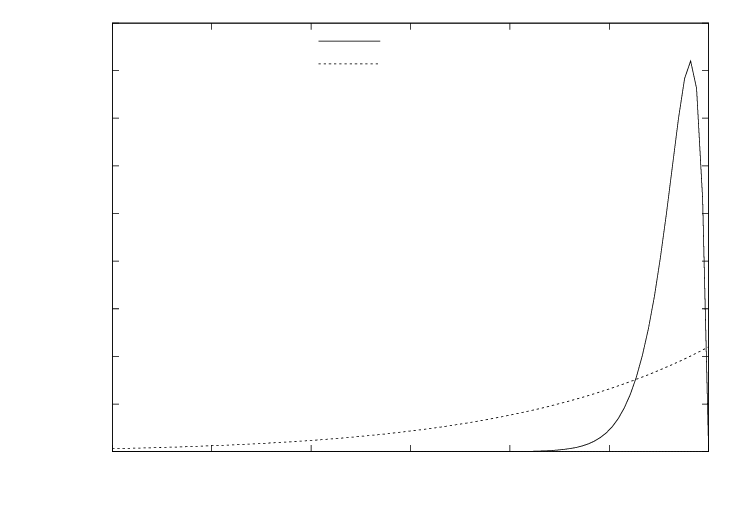}}\gplfronttext
  \end{picture}\endgroup
 	\end{adjustbox}
	\caption{Examples of Beta distributions. $\text{Beta}_{\MLModel_5, 11}(110, 2)$ is retrieved at the end of window $\window_{11}$, $\text{Beta}_{\MLModel_5, 12}(11, 1)$ is used for the next window $\window_{12}$, with  \mem{}$=$$10\%$. \label{fig:beta-memory}}
\end{figure}

\begin{example}\label{ex:process1}
	Following Example~\ref{ex:pr}, let us assume that the current evaluation window $\window_{11}$ in a given court terminates after 200 execution traces according to \dynamicMABVariableWindowMemory. The output of process model assessment is $\{\BetaDistribution{\MLModel_1}{11},$ \ldots$,$ $\BetaDistribution{\MLModel_5}{11}\}$. Figure~\ref{fig:beta-memory} shows $\BetaDistribution{\MLModel_5}{11}$, where $\paralpha{\MLModel_5}{11}$$=$$110$ and $\parbeta{\MLModel_5}{11}$$=$$2$, meaning that $\MLModel_5$ has been frequently sampled and successfully evaluated. Let us then assume that the memory has size $10\%$ (i.e., \mem{}$=$$0.1$). Figure~\ref{fig:beta-memory} shows $\BetaDistribution{\MLModel_5}{12}$ defined for window $\window_{12}$, which is initialized as:
	\begin{enumerate*}
		\item $\paralpha{\MLModel_5}{12}$$=$$110$$\times$$0.1$$=$$11$;
		\item $\parbeta{\MLModel_5}{12}$$=$$2$$\times$$0.1$$=$$0.2$, which is then set to $1$ according to Definition~\ref{def:mem}.
	\end{enumerate*} 
\end{example}
 \section{Model Substitution}\label{sec:substitution}

Process model substitution is executed on the basis of process model assessment in Section~\ref{sec:DMAB}. It takes as input the results of the \dynamicMABVariableWindowMemory\ evaluation in the current window, and returns as output the model to be selected and used by the application in the following window.

\subsection{Ranking-Based Substitution}\label{subsec:substitution-ranking}

Ranking\-/based substitution ranks models $\MLModel$$\in$\candidateList\ in a given window $\window_j$ and determines the model \selected{\MLModel} to be used in the following window $\window_{j+1}$. Let us recall that $\paralpha{\MLModel}{}$ ($\parbeta{\MLModel}{}$, resp.) is incremented by $1$ when $\pij{}.\score{}$ is successfully (unsuccessfully, resp.) evaluated on trace \executionTrace$\in$$\window_j$ (Section~\ref{subsec:buildingblocks-staticmab}). Ranking\-/based substitution is based on a metric evaluating how frequently each model is selected by Thompson Sampling and successfully evaluated in \dynamicMABVariableWindowMemory, as follows.

\begin{definition}[Ranking Metric]\label{def:metric-ranking}
    Let $\window_j$ be a window and \MLModel\ a model. The value of ranking metric $\rankingMetric{\MLModel}{j}$ of $\MLModel$ in $\window_j$ is retrieved as $\paralpha{\MLModel}{j}/(\paralpha{\MLModel}{j}+\parbeta{\MLModel}{j})$.
\end{definition}

According to Definition~\ref{def:metric-ranking}, $\rankingMetric{\MLModel}{j}$ is the ratio between the number of successful evaluations of \MLModel\ (in terms of $\pij{}.\score{}$) and the total number of draws computed by \dynamicMABVariableWindowMemory\ in $\window_j$. It is retrieved for every model in \candidateList\ and used for ranking. 

\vspace{.5em}

\noindent \textbf{Substitution.} At the end of window $\window_j$, the top\-/ranked model \selected{\MLModel} is selected and used within window $\window_{j+1}$. The substitution happens when $\window_j$ terminates according to the value remaining in the experiment (Section~\ref{subsec:DMAB-W}).

We note that the ranking can also be used in case further substitutions in $\window_{j+1}$ are needed. For instance, the second model in the ranking is used when \selected{\MLModel} experiences an (unrecoverable) error.

\begin{example}\label{ex:process2-1}
    Following Example~\ref{ex:process1}, the ranking metric $\rankingMetric{\MLModel_5}{{11}}$ has value $110/(110+2)$$\approx$$0.98$. Let us assume that $\rankingMetric{\MLModel_5}{{11}}$ has the highest value: $\MLModel_5$ substitutes the model used in $\window_{11}$, and is used for bail prediction in $\window_{12}$ (i.e., \selected{\MLModel}$=$$\MLModel_5$).
\end{example}

The assumption that the ranking computed for $\window_j$ is appropriate for $\window_{j+1}$ does not hold when transient changes in the models non-functional behavior are observed within $\window_{j+1}$ (e.g., a sharp change in the environmental context). In this scenario, although \selected{\MLModel} becomes suboptimal, it cannot be substituted until the following window begins. To address this issue, we propose an approach based on early substitution that is presented in Section~\ref{subsec:substitution-assurance}.

\subsection{Assurance-Based Substitution}\label{subsec:substitution-assurance}

 Assurance\-/based substitution triggers early substitution of the selected model \selected{\MLModel} before window \window\ terminates. It monitors \selected{\MLModel} by computing its \emph{assurance level} as follows.

\begin{definition}[Assurance level]\label{def:metric-assurance}
    Let \selected{\MLModel} be the selected model and $\executionTrace_\instant$$\in$$\window_j$ an execution trace. The \emph{assurance level} \assuranceLevel{\instant} of \selected{\MLModel} given $\executionTrace_\instant$ is $\reward_{\selected{\MLModel}^\instant}(\theta) / \reward_\ast (\theta^{(g)})$. 
\end{definition}

According to Definition~\ref{def:metric-assurance}, \assuranceLevel{\instant} is the ratio between \emph{i)} the reward $\reward_{\selected{\MLModel}^\instant}(\theta)$ of the selected model \selected{\MLModel} retrieved at execution trace $\executionTrace_\instant$  and \emph{ii)} the reward $\reward_\ast (\theta^{(g)})$ of the optimal model \optimum{\MLModel}, according to the Monte Carlo simulation in \dynamicMABVariableWindowMemory\ (Section~\ref{subsec:DMAB-W}).
We note that the assurance level can be retrieved for each model $\MLModel_i$ using the corresponding reward as numerator.

The assurance level \assuranceLevelSymbol\ is used to calculate the degradation of the selected model.
 Formally, let $\executionTrace_\instant$ be an execution trace in window $\window_j$.  The \emph{degradation of \selected{\MLModel} at $\executionTrace_\instant$$\in$$\window_j$} is defined as follows.

\begin{equation}
    \degradation{\instant} = 1 - \frac{\sum_{i=1}^\instant \assuranceLevel{i}}{\instant}\label{eq:assurance-degradation}\end{equation}

\vspace{.5em}

\noindent \textbf{Substitution.} It works as the ranking\-/based substitution but the selected model \selected{\MLModel} is substituted with the second model in the ranking before the window termination (i.e., early substitution), \emph{iff} its degradation \degradation{\instant} exceeds threshold \thresholdAssurance\ ($\degradation{\instant}$$>$$\thresholdAssurance$). 

Early substitution copes with transient changes within the window  according to the degradation represented in \thresholdAssurance. A high (low, resp.) threshold means high (low, resp.) tolerance. For instance, a high tolerance is preferable when the substitution overhead is high (e.g., when large models should be physically moved). A low tolerance is preferable when small variations in the properties of the deployed models has a strong impact on the application behavior. We note that, given its fundamental role in the substitution process, we experimentally evaluated the adoption of different degradation thresholds \thresholdAssurance\ in Section~\ref{subsec:experiments-substituion}.

\begin{figure}[t!]
    \centering
    \begin{adjustbox}{max totalsize={.95\columnwidth}{\textheight},center}
        \begingroup
  \makeatletter
  \providecommand\color[2][]{\GenericError{(gnuplot) \space\space\space\@spaces}{Package color not loaded in conjunction with
      terminal option `colourtext'}{See the gnuplot documentation for explanation.}{Either use 'blacktext' in gnuplot or load the package
      color.sty in LaTeX.}\renewcommand\color[2][]{}}\providecommand\includegraphics[2][]{\GenericError{(gnuplot) \space\space\space\@spaces}{Package graphicx or graphics not loaded}{See the gnuplot documentation for explanation.}{The gnuplot epslatex terminal needs graphicx.sty or graphics.sty.}\renewcommand\includegraphics[2][]{}}\providecommand\rotatebox[2]{#2}\@ifundefined{ifGPcolor}{\newif\ifGPcolor
    \GPcolorfalse
  }{}\@ifundefined{ifGPblacktext}{\newif\ifGPblacktext
    \GPblacktexttrue
  }{}\let\gplgaddtomacro\g@addto@macro
\gdef\gplbacktext{}\gdef\gplfronttext{}\makeatother
  \ifGPblacktext
\def\colorrgb#1{}\def\colorgray#1{}\else
\ifGPcolor
      \def\colorrgb#1{\color[rgb]{#1}}\def\colorgray#1{\color[gray]{#1}}\expandafter\def\csname LTw\endcsname{\color{white}}\expandafter\def\csname LTb\endcsname{\color{black}}\expandafter\def\csname LTa\endcsname{\color{black}}\expandafter\def\csname LT0\endcsname{\color[rgb]{1,0,0}}\expandafter\def\csname LT1\endcsname{\color[rgb]{0,1,0}}\expandafter\def\csname LT2\endcsname{\color[rgb]{0,0,1}}\expandafter\def\csname LT3\endcsname{\color[rgb]{1,0,1}}\expandafter\def\csname LT4\endcsname{\color[rgb]{0,1,1}}\expandafter\def\csname LT5\endcsname{\color[rgb]{1,1,0}}\expandafter\def\csname LT6\endcsname{\color[rgb]{0,0,0}}\expandafter\def\csname LT7\endcsname{\color[rgb]{1,0.3,0}}\expandafter\def\csname LT8\endcsname{\color[rgb]{0.5,0.5,0.5}}\else
\def\colorrgb#1{\color{black}}\def\colorgray#1{\color[gray]{#1}}\expandafter\def\csname LTw\endcsname{\color{white}}\expandafter\def\csname LTb\endcsname{\color{black}}\expandafter\def\csname LTa\endcsname{\color{black}}\expandafter\def\csname LT0\endcsname{\color{black}}\expandafter\def\csname LT1\endcsname{\color{black}}\expandafter\def\csname LT2\endcsname{\color{black}}\expandafter\def\csname LT3\endcsname{\color{black}}\expandafter\def\csname LT4\endcsname{\color{black}}\expandafter\def\csname LT5\endcsname{\color{black}}\expandafter\def\csname LT6\endcsname{\color{black}}\expandafter\def\csname LT7\endcsname{\color{black}}\expandafter\def\csname LT8\endcsname{\color{black}}\fi
  \fi
    \setlength{\unitlength}{0.0500bp}\ifx\gptboxheight\undefined \newlength{\gptboxheight}\newlength{\gptboxwidth}\newsavebox{\gptboxtext}\fi \setlength{\fboxrule}{0.5pt}\setlength{\fboxsep}{1pt}\definecolor{tbcol}{rgb}{1,1,1}\begin{picture}(7200.00,5040.00)\gplgaddtomacro\gplbacktext{\csname LTb\endcsname \put(1078,961){\makebox(0,0)[r]{\strut{}$0.992$}}\put(1078,1476){\makebox(0,0)[r]{\strut{}$0.994$}}\put(1078,1990){\makebox(0,0)[r]{\strut{}$0.996$}}\put(1078,2504){\makebox(0,0)[r]{\strut{}$0.998$}}\put(1078,3019){\makebox(0,0)[r]{\strut{}$1$}}\put(1078,3533){\makebox(0,0)[r]{\strut{}$1.002$}}\put(1078,4047){\makebox(0,0)[r]{\strut{}$1.004$}}\put(1078,4562){\makebox(0,0)[r]{\strut{}$1.006$}}\put(1210,484){\makebox(0,0){\strut{}$0$}}\put(2000,484){\makebox(0,0){\strut{}$50$}}\put(2790,484){\makebox(0,0){\strut{}$100$}}\put(3579,484){\makebox(0,0){\strut{}$150$}}\put(4369,484){\makebox(0,0){\strut{}$200$}}\put(5159,484){\makebox(0,0){\strut{}$250$}}\put(5923,961){\makebox(0,0)[l]{\strut{}$0.84$}}\put(5923,1476){\makebox(0,0)[l]{\strut{}$0.86$}}\put(5923,1990){\makebox(0,0)[l]{\strut{}$0.88$}}\put(5923,2504){\makebox(0,0)[l]{\strut{}$0.9$}}\put(5923,3019){\makebox(0,0)[l]{\strut{}$0.92$}}\put(5923,3533){\makebox(0,0)[l]{\strut{}$0.94$}}\put(5923,4047){\makebox(0,0)[l]{\strut{}$0.96$}}\put(5923,4562){\makebox(0,0)[l]{\strut{}$0.98$}}}\gplgaddtomacro\gplfronttext{\csname LTb\endcsname \put(4804,1757){\makebox(0,0)[r]{\strut{}$\MLModel_5$}}\csname LTb\endcsname \put(4804,1537){\makebox(0,0)[r]{\strut{}$\MLModel_4$}}\csname LTb\endcsname \put(4804,1317){\makebox(0,0)[r]{\strut{}trend line $\MLModel_5$}}\csname LTb\endcsname \put(4804,1097){\makebox(0,0)[r]{\strut{}trend line $\MLModel_4$}}\csname LTb\endcsname \put(4804,877){\makebox(0,0)[r]{\strut{}value remaining}}\csname LTb\endcsname \put(209,2761){\rotatebox{-270.00}{\makebox(0,0){\strut{}Assurance level \assuranceLevelSymbol}}}\put(6693,2761){\rotatebox{-270.00}{\makebox(0,0){\strut{}Value remaining}}}\put(3500,154){\makebox(0,0){\strut{}Execution traces \executionTrace}}}\gplbacktext
    \put(0,0){\includegraphics[width={360.00bp},height={252.00bp}]{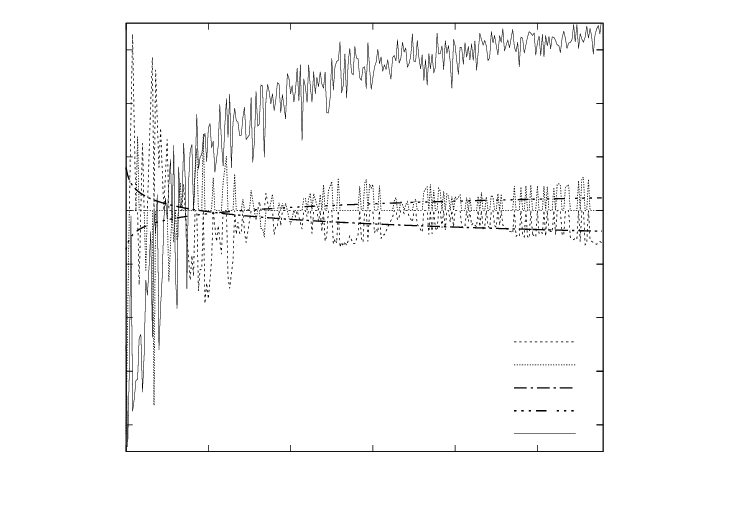}}\gplfronttext
  \end{picture}\endgroup
     \end{adjustbox}
    \caption{Assurance levels of $\selected{\MLModel_5}$ (ranked first) and $\MLModel_4$ (ranked second), denoted as $\assuranceLevel{\selected{\MLModel_5}, \instant}$ and $\assuranceLevel{\MLModel_4, \instant}$, respectively.  
    The plot shows the logarithmic trend lines and the outcomes in relation to the value remaining in the experiment of the \dynamicMABVariableWindowMemory\ in a given time window \window. \label{fig:vandass}}
\end{figure}

\begin{example}\label{ex:process2-2}
    Following Example~\ref{ex:process2-1}, let us consider model $\MLModel_5$ as the selected model and model $\MLModel_4$ as the second model in the ranking. Figure~\ref{fig:vandass} shows an example of the assurance levels of $\selected{\MLModel_5}$ and $\MLModel_4$, denoted as $\assuranceLevel{\selected{\MLModel_5}, \instant}$ and $\assuranceLevel{\MLModel_4, \instant}$, respectively. Figure~\ref{fig:vandass} also shows the corresponding logarithmic trend lines for readability, and the value remaining in the experiment, using \dynamicMABVariableWindowMemory.
    
    Let us first consider ranking\-/based substitution only. At \instant$=$$290$, window  $\window_j$ terminates according to the value remaining in the experiment (Section~\ref{subsec:DMAB-W}). \dynamicMABVariableWindowMemory\ recomputes the ranking; $\MLModel_4$ is the top\-/ranked model while $\MLModel_5$ the second one. \dynamicMABVariableWindowMemory\ triggers ranking\-/based substitution, and $\MLModel_4$ becomes the selected model (\selected{\MLModel}$=$$\MLModel_4$) for window $\window_{j+1}$.

    Let us then consider assurance\-/based substitution. 
    We can observe that $\assuranceLevel{\selected{\MLModel_5}, \instant}$ decreases as execution traces arrive. From \instant$=$$38$, $\assuranceLevel{\selected{\MLModel_5}, \instant}$ stably becomes less than $1$. Around \instant$=$$87$, $\assuranceLevel{\selected{\MLModel_4}, \instant}$ overcomes $\assuranceLevel{\selected{\MLModel_5}, \instant}$ thus suggesting a possible substitution. 
    However, the degradation  of model $\MLModel_5$ is not severe enough to justify the early substitution (i.e., the degradation is lower than the degradation threshold). 
\end{example}

\section{Walkthrough}\label{sec:walkthrough}

We present a walkthrough of our approach based on the reference scenario in Example~\ref{ex:scenario}. Figure~\ref{fig:pseudocode} shows the pseudocode of our approach.  

The five models in the candidate list \candidateList\ in Example~\ref{ex:scenario} are first evaluated offline using the Static MAB in Section~\ref{subsec:buildingblocks-staticmab}, to retrieve the optimum model \optimum{\MLModel} that initializes our approach. Let us assume that model $\MLModel_2$ is selected as the optimal model (\selected{\MLModel}$=$\optimum{\MLModel}).
Our model assessment and substitution processes (Figure~\ref{fig:pseudocode}) begin, instantiating the \dynamicMABVariableWindowMemory. The processes take as input \begin{enumerate*}
	\item the models in \candidateList, \item the observed execution traces, \item the non\-/functional property fairness in Example~\ref{ex:pr}, \item the memory size \mem{}, \item the early substitution threshold \thresholdAssurance, and \item the minimum number of MAB iterations.
\end{enumerate*} 
Execution traces observed from all the models are given as input to process model assessment. For each execution trace, the function \funcName{thompson\_sampling} in \dynamicMABVariableWindowMemory\ chooses a model among those in \candidateList\ by drawing a sample from each model Beta distribution and retrieving the one with the highest value. The retrieved model is evaluated according to the fairness score function (function \funcName{score\_function} in Figure~\ref{fig:pseudocode-fairness}), updating the corresponding Beta distribution accordingly (Definition~\ref{def:mem}). Then, process model assessment invokes function \funcName{monte\_carlo\_simulation} to simulate the probabilities of models being winners. It creates a two\-/dimensional matrix with dimensions $\vert \candidateList \vert$$\times$$g$, where $g$ is the number of estimations. Each cell contains samples drawn from the models' Beta distributions; the matrix counts the frequency of each model being winner and approximates the probability distribution \pwinner, accordingly. 

Process model assessment proceeds until the minimum number of iterations is met and the value remaining in the experiment permits to reach a statistically relevant decision (function \funcName{should\_terminate}). At this point, the evaluation window ends (function \funcName{handle\_window}).

Upon process model assessment ends, process model substitution invokes function \funcName{send\_into\_production} ranking models proportionally to the number of their successful evaluations of the non\-/functional property (ranking metric in Definition~\ref{def:metric-ranking}). For instance, the ranking at the end of window $\window_1$ is $\{\MLModel_3,$ $\MLModel_2,$ $\MLModel_1,$ $\MLModel_4,$ $\MLModel_5\}$, from best to worst. The top\-/ranked model ($\MLModel_3$) is pushed to production replacing $\MLModel_2$ selected at deployment time by the Static MAB (Section~\ref{subsec:substitution-ranking}). 
Process model substitution also monitors the selected model invoking function \funcName{assurance\_management}. The latter verifies whether the non\-/functional behavior of the selected model is worsening with respect to the optimum model estimated by the Monte Carlo simulation. It computes the assurance level \assuranceLevel{\instant} for each new execution trace \executionTrace$_\instant$ (Definition~\ref{def:metric-assurance}) and uses it to retrieve the overall degradation (equation~\ref{eq:assurance-degradation}).
For instance, during window $\window_2$, the degradation of $\MLModel_3$ is negligible, meaning that $\MLModel_3$ is still adequate according to the data observed in $\window_2$ and does not need to be  substituted in advance.

When the current window $\window_2$ terminates, process model substitution recomputes the ranking. For instance, the ranking is $\{\MLModel_3, \MLModel_2, \MLModel_4, \MLModel_1, \MLModel_5\}$, and $\MLModel_3$ is used as the selected model for window $\window_3$. In $\window_3$, process model substitution observes a constant degradation in the assurance level of $\MLModel_3$, reaching the early substitution threshold. Early substitution is therefore triggered and the second model in the ranking ($\MLModel_2$) substitutes $\MLModel_3$. Again when the current window $\window_3$ terminates, process model substitution recomputes the ranking and $\MLModel_2$ is confirmed at the top of the ranking. 

Overall, this adaptive approach ensures that \begin{enumerate*}
	\item model substitution happens only when the decision is statistically relevant according to the observed behavior (\emph{ranking\-/based substitution} in \dynamicMABVariableWindowMemory),
	\item a sub\-/optimal substitution decision can be fixed as soon as it is detected without waiting for the entire evaluation window (\emph{assurance\-/based substitution}), and
	\item the entire process can be fine\-/tuned according to each scenario.
\end{enumerate*}

\begin{figure*}[!t]
	\begin{adjustbox}{max totalsize={.95\textwidth}{\textheight},center}
		\begin{footnotesize}
        \hrulefill\vspace{3pt}
        \begin{tabular}{ccc}
            \hspace{-.2cm}
            \parbox{6.cm}{
                \begin{tabbing}
                
                    \INPUT\\
                        \candidateList: models to be ranked \\ 
                        \executionTrace[]: execution traces \\
                        \pij{}: target non-functional property\\
                        \mem{}: memory size\\
                        \thresholdAssurance: early substitution threshold\\
                        \burnin:  minimum number of iterations\\
                        ~\\[1pt]
                    \MAIN\\
                        $i$$=$$0$ \commentall{ Iteration counter }\\
                        \\
                        \commentall{ Begin the DMVW-Mem } \\
                        \com{for}\=\ \com{each} $\executionTrace_\instant$$\in$\executionTrace[]\\
                            \uno \optimum{\MLModel}$=$\thompsonsampm(\candidateList)\\
\uno \com{apply\_score\_function}(\=$\executionTrace_t$, \optimum{\MLModel},\\ \due \VARIANCETHRESHOLD)\\
                            \uno (\mc[], \pwinner)$=$\mcsimulation (\candidateList)\\
                            \uno \estmodels[]$=$record current estimates of\\
                            \uno each \MLModel$\in$\candidateList\ wins according to ranking metric\\ \uno \com{if}\=\ ($i$$>$\burnin)\=\ \com{AND} \shouldterminate(\pwinner, \\
                            \tre \estmodels, \mc[])\\ \due \handlewindow()\\
\uno \com{else}\\
                            \due $i$$=$$i$$+$$1$\\
                            \due \commentall{ Continue with the next trace } \\

                \end{tabbing}
            }&
            \parbox{.325\textwidth}{
                \begin{tabbing}

                    \textbf{THOMPSON\_SAMPLING}\\
                        \com{for}\=\ \com{each} \MLModel$\in$\candidateList\\
                            \uno \sample$=$draw sample from \BetaDistribution{\MLModel}{}\\
                        \optimum{\MLModel}$=$$\arg\max$ \sample \\
                        \return\ \optimum{\MLModel} \\ 
                    ~\\[1pt] 

                    \textbf{MONTE\_CARLO\_SIMULATION}\\
                        $g$$=$$100$\\
                        \mc[]$=$\=\ matrix $\vert\candidateList\vert$$\times$$g$ of samples \\
                            \uno drawn from \BetaDistribution{\MLModel}{} $\forall$\MLModel$\in$\candidateList\\
                        compute \ensuremath{p(\theta\mid y_\instant)} from \mc[]\\
                        \return\ \mc[], \pwinner\\ 	
                    ~\\[1pt]
                    
                    \textbf{SHOULD\_TERMINATE}\\
                        \thresholdRegret$=$$0.01$\\
                        compute $\reward_{\optimumTime{\MLModel}{\instant}}(\theta_0)$ from \estmodels[]\\
                        $\regret^{(g)}$$=$$\frac{\reward_{\ast}(\theta^{(g)})-\reward_{\optimumTime{\MLModel}{\instant}}(\theta^{(g)})}{\reward_{\optimumTime{\MLModel}{\instant}}(\theta^{(g)})}$\\
                        \return\ $\text{\emph{percentile}}$\=$(\regret^{(g)}, 0.95)$$\leq$$\reward_{\optimumTime{\MLModel}{\instant}}(\theta_0)\times$\\
                        \uno \thresholdRegret\\
                    ~\\[1pt]

                    \textbf{HANDLE\_WINDOW}\\
                        \productionm(\selected{\MLModel}, \candidateList)\\
$j-1$, $j$ $=$ previous and current windows\\
                        \com{for}\=\ \com{each} \MLModel$\in$\candidateList \\
                            \uno \commentall{ Initialize Beta distrib. according to \mem{} }\\
                            \uno \paralpha{\MLModel}{j}$=$\paralpha{\MLModel}{j-1}$\times$\mem{}\\
                            \uno \parbeta{\MLModel}{j}$=$\parbeta{\MLModel}{j-1}$\times$\mem{}\\   

                \end{tabbing}
            }&
            \parbox{6.cm}{
                \begin{tabbing}

                    \textbf{SEND\_INTO\_PRODUCTION}\\
\ranking$=$\=sort \candidateList\ in descending order\\
                            \uno according to $\paralpha{\MLModel}{}/(\paralpha{\MLModel}{}+\parbeta{\MLModel}{})$\\
                        \assurancemgmntm(\ranking)\\ \selected{\MLModel}$=$\ranking[0]\\
                        \push{\selected{\MLModel}} into production \\
                    ~\\[1pt]

                    \textbf{ASSURANCE\_MANAGEMENT}\\
                    \cumulative$=$$0$\\
                    $i$$=$$1$ \commentall{ counter }\\
                    \com{for}\=\ \com{each} $\executionTrace_\instant$$\in$\window \\
                        \uno \assuranceLevel{\instant}$=$$\reward_{\selected{\MLModel}^\instant}(\theta) / \reward_\ast (\theta^{(g)})$\\
                        \uno \cumulative$=$\cumulative$+$\assuranceLevel{\instant}\\
                        \uno \degradation{\instant}$=$$1-\cumulative/i$\\
                        \uno \com{if} \=\ \degradation{\instant}$>$\thresholdAssurance\\
                            \due \commentall{ Early substitution }\\
                            \due \selected{\MLModel}$=$$\ranking[1]$\\
                            \due \push{\selected{\MLModel}} into production \\
\uno $i$$=$$i$$+$$1$\\
                \end{tabbing}
            }
        \end{tabular}
        \hrulefill
        \vspace{10pt} 

    \end{footnotesize}
 	\end{adjustbox}
	\caption{Pseudocode of our approach. \label{fig:pseudocode}}
\end{figure*}
 \section{Experimental Evaluation}\label{sec:experiments}
We experimentally evaluated our approach focusing on: \emph{i)} the model assessment at development time using Static MAB; \emph{ii)} the model substitution at run time using Dynamic MAB, also evaluating the impact of different memory sizes;  \emph{iii)} quality and \emph{iv)} performance of ranking\-/based and assurance\-/based substitutions.

\subsection{Experimental Settings}\label{subsec:experiments-setting}

We considered the application for bail estimation and property fairness in our reference example in Section~\ref{sec:walkthrough}.
In our experiments, we used the dataset of the Connecticut State Department of Correction.\footnote{Available at \url{https://data.ct.gov/Public-Safety/Accused-Pre-Trial-Inmates-in-Correctional-Faciliti/b674-jy6w} and downloaded on February 21\textsuperscript{st}, 2020.} This dataset provides a daily updated list of people detained in the Department's facilities awaiting a trial. It anonymously discloses data of individual people detained in the correctional facilities every day starting from July 1\textsuperscript{st}, 2016. It contains attributes such as last admission date, race, gender, age, type of offence and facility description, in more than four millions data points (at the download date). We divided this set into training and test sets, where the training set includes more than 3 million points. 

 We modeled the score function $\pij{}.\score{}$ of property fairness as the variance ($\sigma^2$) of the bail amount in relation to sensitive attributes \emph{gender} and \emph{race}~\cite{CAPAI, 10.1007/978-3-031-46846-91, 10479352}. Figure~\ref{fig:pseudocode-fairness} shows the pseudocode of the score function and its usage according to the threshold\-/based evaluation in Section~\ref{subsec:substitution-assurance}. 
We generated five Naive Bayes models \candidateList$=$$\{\MLModel_1,$ \ldots$,$ $\MLModel_{5}\}$, each one trained on a training set randomly extracted from the main training set. The models showed similar performance, in terms of precision and recall in bail estimation.
We also extracted 10 test sets corresponding to 10 individual experiments $\experiment{1}$--$\experiment{10}$ to be used in our experimental evaluation. 

\begin{figure}[!t]

\begin{footnotesize}
    \centering
    \begin{tabbing}
    \INPUT\\ 
        \executionTrace: execution trace\\
        \optimum{\MLModel}: Thompson selected model\\ 
        \VARIANCETHRESHOLD:  variance threshold\\
        ~\\[1pt]
    \OUTPUT\\
        \paralpha{\optimum{\MLModel}}{} and \parbeta{\optimum{\MLModel}}{}\\

        ~\\[1pt]
\textbf{SCORE\_FUNCTION}\\
        \rows[] = \executionTrace \\
        \rows[] += \= generate test data for all protected\\ 
        \uno groups against \executionTrace \\
        \predictions[] = \optimum{\MLModel}.predict(\rows[])\\
        \variancem$=$variance(\predictions[])\\
        \return\ \variancem\\
        ~\\[1pt]

    \textbf{APPLY\_SCORE\_FUNCTION}\\
        \variancem$=$\textbf{score\_function}(\executionTrace)\\
        \com{if}\=\ (\variancem$<$\VARIANCETHRESHOLD)\\
        \uno \paralpha{\optimum{\MLModel}}{}$=$\paralpha{\optimum{\MLModel}}{}$+$$1$\\
        \com{else}\\
        \uno \parbeta{\optimum{\MLModel}}{}$=$\parbeta{\optimum{\MLModel}}{}$+$$1$
\end{tabbing}
\end{footnotesize}

 \caption{Pseudocode of the score function of property fairness and its usage. \label{fig:pseudocode-fairness}}
\end{figure}

Experiments have been run on a laptop running Microsoft Windows 10, equipped with a CPU Intel Core i7 @ 2.6 GHz and 16 GBs of RAM, using Python 3 with libraries \emph{numpy} v1.19.1~\cite{harris2020array}, \emph{pandas} v1.2.5~\cite{mckinney-proc-scipy-2010,reback2020pandas} and \emph{scikit-learn} v0.22.1~\cite{scikit-learn}. Datasets, code, and experimental results are available at \url{https://doi.org/10.13130/RD_UNIMI/2G3CVO}.

\subsection{Model Assessment}\label{subsec:experiments-models}
We present the experimental evaluation of our Static MAB for model assessment at development time. We compare the five Naive Bayes models using the Static MAB approach, by evaluating their behavior with respect to non-functional property fairness. 
Table~\ref{tab:sMAB} shows the Thompson Sampling draws for the five models in the candidate list on a randomly chosen sample (2,000 data points) for each of the 10 experiments.

\begin{table}[!t]
	\centering
	\caption{Static MAB comparison in terms of Thompson Sampling draws on a random sample of 2,000 data points for each experiment. The best candidate for each experiment is denoted in bold.
\label{tab:sMAB}}
	\footnotesize
	\begin{adjustbox}{max totalsize={.95\columnwidth}{\textheight},center}
		{\footnotesize
\begin{tabular}{c|c|c|c|c|c}
    \toprule
\textbf{Experiments} & \textbf{$\MLModel_1$} & \textbf{$\MLModel_2$} & \textbf{$\MLModel_3$} & \textbf{$\MLModel_4$} & \textbf{$\MLModel_5$} \\
    \midrule
$exp_1$              & \textbf{1,021}        & 156                   & 609                   & 28                    & 186                   \\
    $exp_2$              & 4                     & \textbf{876}          & 349                   & 462                   & 309                   \\
    $exp_3$              & 414                   & 286                   & 341                   & \textbf{645}          & 314                   \\
    $exp_4$              & 55                    & 198                   & 670                   & \textbf{1,028}        & 49                    \\
    $exp_5$              & 432                   & 84                    & 208                   & \textbf{666}          & 610                   \\
    $exp_6$              & 138                   & 255                   & 50                    & 607                   & \textbf{950}          \\
    $exp_7$              & 419                   & 31                    & 268                   & 514                   & \textbf{768}          \\
    $exp_8$              & 205                   & 528                   & 135                   & 104                   & \textbf{1,028}        \\
    $exp_9$              & 288                   & 394                   & 35                    & 43                    & \textbf{1,240}        \\
    $exp_{10}$           & 104                   & \textbf{1278}         & 112                   & 453                   & 53                    \\
\bottomrule
\end{tabular}
} 	\end{adjustbox}
\end{table}  

Table~\ref{tab:sMAB} shows the distribution of models selected as best candidate (denoted in bold) for property fairness. Since $\MLModel_3$ is never selected as the best candidate, it is removed from the candidate list for the rest of the experimental evaluation.
We note that comparing models based on the same algorithm (i.e., Naive Bayes) is more challenging than considering different algorithms~\cite{AADP.MEDES2020}, posing our experiments in a worst-case scenario. 

\subsection{Model Substitution}\label{subsec:experiments-substituion}
We present the experimental evaluation of our process model substitution using \dynamicMABVariableWindowMemory\ with different memory sizes ($\mem{0}$$=$$0\%$, \mem{5}$=$$5\%$, \mem{10}$=$$10\%$, \mem{25}$=$$25\%$). 
We evaluated \emph{i)} the impact of the memory on the window size, \emph{ii)} the impact of the ranking-based substitution in terms of stability of model selections, \emph{iii)} the quality of the ranking-based substitution, and \emph{iv)} the quality of the assurance-based substitution. 
We note that no artificial degradation was introduced during the experiments.

\subsubsection{Memory Size and Ranking}\label{subsubsec:experiments-substitution-memory}

\begin{figure}[!t]
	\centering
\begin{adjustbox}{max totalsize={.95\columnwidth}{\textheight},center}
		\begingroup
  \makeatletter
  \providecommand\color[2][]{\GenericError{(gnuplot) \space\space\space\@spaces}{Package color not loaded in conjunction with
      terminal option `colourtext'}{See the gnuplot documentation for explanation.}{Either use 'blacktext' in gnuplot or load the package
      color.sty in LaTeX.}\renewcommand\color[2][]{}}\providecommand\includegraphics[2][]{\GenericError{(gnuplot) \space\space\space\@spaces}{Package graphicx or graphics not loaded}{See the gnuplot documentation for explanation.}{The gnuplot epslatex terminal needs graphicx.sty or graphics.sty.}\renewcommand\includegraphics[2][]{}}\providecommand\rotatebox[2]{#2}\@ifundefined{ifGPcolor}{\newif\ifGPcolor
    \GPcolorfalse
  }{}\@ifundefined{ifGPblacktext}{\newif\ifGPblacktext
    \GPblacktexttrue
  }{}\let\gplgaddtomacro\g@addto@macro
\gdef\gplbacktext{}\gdef\gplfronttext{}\makeatother
  \ifGPblacktext
\def\colorrgb#1{}\def\colorgray#1{}\else
\ifGPcolor
      \def\colorrgb#1{\color[rgb]{#1}}\def\colorgray#1{\color[gray]{#1}}\expandafter\def\csname LTw\endcsname{\color{white}}\expandafter\def\csname LTb\endcsname{\color{black}}\expandafter\def\csname LTa\endcsname{\color{black}}\expandafter\def\csname LT0\endcsname{\color[rgb]{1,0,0}}\expandafter\def\csname LT1\endcsname{\color[rgb]{0,1,0}}\expandafter\def\csname LT2\endcsname{\color[rgb]{0,0,1}}\expandafter\def\csname LT3\endcsname{\color[rgb]{1,0,1}}\expandafter\def\csname LT4\endcsname{\color[rgb]{0,1,1}}\expandafter\def\csname LT5\endcsname{\color[rgb]{1,1,0}}\expandafter\def\csname LT6\endcsname{\color[rgb]{0,0,0}}\expandafter\def\csname LT7\endcsname{\color[rgb]{1,0.3,0}}\expandafter\def\csname LT8\endcsname{\color[rgb]{0.5,0.5,0.5}}\else
\def\colorrgb#1{\color{black}}\def\colorgray#1{\color[gray]{#1}}\expandafter\def\csname LTw\endcsname{\color{white}}\expandafter\def\csname LTb\endcsname{\color{black}}\expandafter\def\csname LTa\endcsname{\color{black}}\expandafter\def\csname LT0\endcsname{\color{black}}\expandafter\def\csname LT1\endcsname{\color{black}}\expandafter\def\csname LT2\endcsname{\color{black}}\expandafter\def\csname LT3\endcsname{\color{black}}\expandafter\def\csname LT4\endcsname{\color{black}}\expandafter\def\csname LT5\endcsname{\color{black}}\expandafter\def\csname LT6\endcsname{\color{black}}\expandafter\def\csname LT7\endcsname{\color{black}}\expandafter\def\csname LT8\endcsname{\color{black}}\fi
  \fi
    \setlength{\unitlength}{0.0500bp}\ifx\gptboxheight\undefined \newlength{\gptboxheight}\newlength{\gptboxwidth}\newsavebox{\gptboxtext}\fi \setlength{\fboxrule}{0.5pt}\setlength{\fboxsep}{1pt}\definecolor{tbcol}{rgb}{1,1,1}\begin{picture}(7200.00,5040.00)\gplgaddtomacro\gplbacktext{\csname LTb\endcsname \put(814,704){\makebox(0,0)[r]{\strut{}$0$}}\put(814,1527){\makebox(0,0)[r]{\strut{}$100$}}\put(814,2350){\makebox(0,0)[r]{\strut{}$200$}}\put(814,3173){\makebox(0,0)[r]{\strut{}$300$}}\put(814,3996){\makebox(0,0)[r]{\strut{}$400$}}\put(814,4819){\makebox(0,0)[r]{\strut{}$500$}}\put(946,484){\makebox(0,0){\strut{}$0$}}\put(2104,484){\makebox(0,0){\strut{}$10000$}}\put(3261,484){\makebox(0,0){\strut{}$20000$}}\put(4419,484){\makebox(0,0){\strut{}$30000$}}\put(5577,484){\makebox(0,0){\strut{}$40000$}}\put(6735,484){\makebox(0,0){\strut{}$50000$}}}\gplgaddtomacro\gplfronttext{\csname LTb\endcsname \put(5816,4646){\makebox(0,0)[r]{\strut{}moving avg \mem{0}}}\put(5816,4426){\makebox(0,0)[r]{\strut{}moving avg \mem{5}}}\put(5816,4206){\makebox(0,0)[r]{\strut{}moving avg \mem{10}}}\put(5816,3986){\makebox(0,0)[r]{\strut{}moving avg \mem{25}}}\put(209,2761){\rotatebox{-270.00}{\makebox(0,0){\strut{}Window size $\vert \window \vert$}}}\put(3874,154){\makebox(0,0){\strut{}Execution traces \executionTrace}}}\gplbacktext
    \put(0,0){\includegraphics[width={360.00bp},height={252.00bp}]{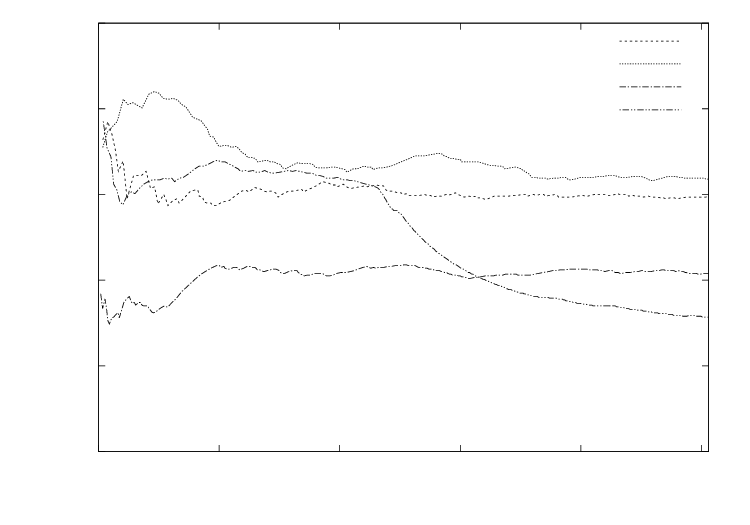}}\gplfronttext
  \end{picture}\endgroup
 	\end{adjustbox}
\caption{Individual window sizes and moving average trends across all sets of execution traces with different memory sizes for $\experiment{1}$.}
	\label{fig:mem}
\end{figure}

Figure~\ref{fig:mem} shows the window size varying the memory in experiment $\experiment{1}$ with residual threshold \thresholdRegret$=$$0.01$ (Section~\ref{subsec:DMAB-W}). We note that a bigger memory corresponds to a smaller window. This is expected, since the \dynamicMABVariableWindowMemory\ does not start from scratch in every window, and the more \dynamicMABVariableWindowMemory\ knows about the models' Beta distributions, the sooner the value remaining in the experiment reaches the threshold. Considering all the experiments, the average window size for \mem{25} is 157 confirming the trend in Figure~\ref{fig:mem}. 

\begin{figure}[!t]
  \centering
	\resizebox{0.49\textwidth}{!}{
	
		\begin{tabular}{c}

			\begin{tabular}{*{2}{c}}

				\begin{adjustbox}{max totalsize={.95\columnwidth}{\textheight},center}
					\begingroup
  \makeatletter
  \providecommand\color[2][]{\GenericError{(gnuplot) \space\space\space\@spaces}{Package color not loaded in conjunction with
      terminal option `colourtext'}{See the gnuplot documentation for explanation.}{Either use 'blacktext' in gnuplot or load the package
      color.sty in LaTeX.}\renewcommand\color[2][]{}}\providecommand\includegraphics[2][]{\GenericError{(gnuplot) \space\space\space\@spaces}{Package graphicx or graphics not loaded}{See the gnuplot documentation for explanation.}{The gnuplot epslatex terminal needs graphicx.sty or graphics.sty.}\renewcommand\includegraphics[2][]{}}\providecommand\rotatebox[2]{#2}\@ifundefined{ifGPcolor}{\newif\ifGPcolor
    \GPcolorfalse
  }{}\@ifundefined{ifGPblacktext}{\newif\ifGPblacktext
    \GPblacktexttrue
  }{}\let\gplgaddtomacro\g@addto@macro
\gdef\gplbacktext{}\gdef\gplfronttext{}\makeatother
  \ifGPblacktext
\def\colorrgb#1{}\def\colorgray#1{}\else
\ifGPcolor
      \def\colorrgb#1{\color[rgb]{#1}}\def\colorgray#1{\color[gray]{#1}}\expandafter\def\csname LTw\endcsname{\color{white}}\expandafter\def\csname LTb\endcsname{\color{black}}\expandafter\def\csname LTa\endcsname{\color{black}}\expandafter\def\csname LT0\endcsname{\color[rgb]{1,0,0}}\expandafter\def\csname LT1\endcsname{\color[rgb]{0,1,0}}\expandafter\def\csname LT2\endcsname{\color[rgb]{0,0,1}}\expandafter\def\csname LT3\endcsname{\color[rgb]{1,0,1}}\expandafter\def\csname LT4\endcsname{\color[rgb]{0,1,1}}\expandafter\def\csname LT5\endcsname{\color[rgb]{1,1,0}}\expandafter\def\csname LT6\endcsname{\color[rgb]{0,0,0}}\expandafter\def\csname LT7\endcsname{\color[rgb]{1,0.3,0}}\expandafter\def\csname LT8\endcsname{\color[rgb]{0.5,0.5,0.5}}\else
\def\colorrgb#1{\color{black}}\def\colorgray#1{\color[gray]{#1}}\expandafter\def\csname LTw\endcsname{\color{white}}\expandafter\def\csname LTb\endcsname{\color{black}}\expandafter\def\csname LTa\endcsname{\color{black}}\expandafter\def\csname LT0\endcsname{\color{black}}\expandafter\def\csname LT1\endcsname{\color{black}}\expandafter\def\csname LT2\endcsname{\color{black}}\expandafter\def\csname LT3\endcsname{\color{black}}\expandafter\def\csname LT4\endcsname{\color{black}}\expandafter\def\csname LT5\endcsname{\color{black}}\expandafter\def\csname LT6\endcsname{\color{black}}\expandafter\def\csname LT7\endcsname{\color{black}}\expandafter\def\csname LT8\endcsname{\color{black}}\fi
  \fi
    \setlength{\unitlength}{0.0500bp}\ifx\gptboxheight\undefined \newlength{\gptboxheight}\newlength{\gptboxwidth}\newsavebox{\gptboxtext}\fi \setlength{\fboxrule}{0.5pt}\setlength{\fboxsep}{1pt}\definecolor{tbcol}{rgb}{1,1,1}\begin{picture}(7200.00,5040.00)\gplgaddtomacro\gplbacktext{\csname LTb\endcsname \put(660,1431){\makebox(0,0)[r]{\strut{}\Large $\MLModel_1$}}\put(660,2158){\makebox(0,0)[r]{\strut{}\Large $\MLModel_2$}}\put(660,2885){\makebox(0,0)[r]{\strut{}\Large $\MLModel_4$}}\put(660,3612){\makebox(0,0)[r]{\strut{}\Large $\MLModel_5$}}\put(792,484){\makebox(0,0){\strut{}$0$}}\put(1971,484){\makebox(0,0){\strut{}$10000$}}\put(3149,484){\makebox(0,0){\strut{}$20000$}}\put(4328,484){\makebox(0,0){\strut{}$30000$}}\put(5507,484){\makebox(0,0){\strut{}$40000$}}\put(6685,484){\makebox(0,0){\strut{}$50000$}}}\gplgaddtomacro\gplfronttext{\csname LTb\endcsname \put(-77,2521){\rotatebox{-270.00}{\makebox(0,0){\strut{}\Large Models}}}\put(3797,154){\makebox(0,0){\strut{}\Large Execution traces \executionTrace}}\put(3797,4709){\makebox(0,0){\strut{}\Large \mem{0}}}}\gplbacktext
    \put(0,0){\includegraphics[width={360.00bp},height={252.00bp}]{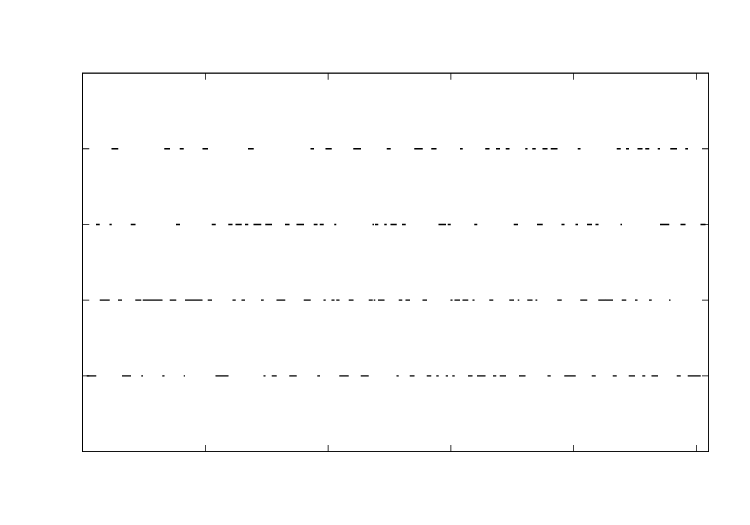}}\gplfronttext
  \end{picture}\endgroup
 				\end{adjustbox} &
				\begin{adjustbox}{max totalsize={.95\columnwidth}{\textheight},center}
					\begingroup
  \makeatletter
  \providecommand\color[2][]{\GenericError{(gnuplot) \space\space\space\@spaces}{Package color not loaded in conjunction with
      terminal option `colourtext'}{See the gnuplot documentation for explanation.}{Either use 'blacktext' in gnuplot or load the package
      color.sty in LaTeX.}\renewcommand\color[2][]{}}\providecommand\includegraphics[2][]{\GenericError{(gnuplot) \space\space\space\@spaces}{Package graphicx or graphics not loaded}{See the gnuplot documentation for explanation.}{The gnuplot epslatex terminal needs graphicx.sty or graphics.sty.}\renewcommand\includegraphics[2][]{}}\providecommand\rotatebox[2]{#2}\@ifundefined{ifGPcolor}{\newif\ifGPcolor
    \GPcolorfalse
  }{}\@ifundefined{ifGPblacktext}{\newif\ifGPblacktext
    \GPblacktexttrue
  }{}\let\gplgaddtomacro\g@addto@macro
\gdef\gplbacktext{}\gdef\gplfronttext{}\makeatother
  \ifGPblacktext
\def\colorrgb#1{}\def\colorgray#1{}\else
\ifGPcolor
      \def\colorrgb#1{\color[rgb]{#1}}\def\colorgray#1{\color[gray]{#1}}\expandafter\def\csname LTw\endcsname{\color{white}}\expandafter\def\csname LTb\endcsname{\color{black}}\expandafter\def\csname LTa\endcsname{\color{black}}\expandafter\def\csname LT0\endcsname{\color[rgb]{1,0,0}}\expandafter\def\csname LT1\endcsname{\color[rgb]{0,1,0}}\expandafter\def\csname LT2\endcsname{\color[rgb]{0,0,1}}\expandafter\def\csname LT3\endcsname{\color[rgb]{1,0,1}}\expandafter\def\csname LT4\endcsname{\color[rgb]{0,1,1}}\expandafter\def\csname LT5\endcsname{\color[rgb]{1,1,0}}\expandafter\def\csname LT6\endcsname{\color[rgb]{0,0,0}}\expandafter\def\csname LT7\endcsname{\color[rgb]{1,0.3,0}}\expandafter\def\csname LT8\endcsname{\color[rgb]{0.5,0.5,0.5}}\else
\def\colorrgb#1{\color{black}}\def\colorgray#1{\color[gray]{#1}}\expandafter\def\csname LTw\endcsname{\color{white}}\expandafter\def\csname LTb\endcsname{\color{black}}\expandafter\def\csname LTa\endcsname{\color{black}}\expandafter\def\csname LT0\endcsname{\color{black}}\expandafter\def\csname LT1\endcsname{\color{black}}\expandafter\def\csname LT2\endcsname{\color{black}}\expandafter\def\csname LT3\endcsname{\color{black}}\expandafter\def\csname LT4\endcsname{\color{black}}\expandafter\def\csname LT5\endcsname{\color{black}}\expandafter\def\csname LT6\endcsname{\color{black}}\expandafter\def\csname LT7\endcsname{\color{black}}\expandafter\def\csname LT8\endcsname{\color{black}}\fi
  \fi
    \setlength{\unitlength}{0.0500bp}\ifx\gptboxheight\undefined \newlength{\gptboxheight}\newlength{\gptboxwidth}\newsavebox{\gptboxtext}\fi \setlength{\fboxrule}{0.5pt}\setlength{\fboxsep}{1pt}\definecolor{tbcol}{rgb}{1,1,1}\begin{picture}(7200.00,5040.00)\gplgaddtomacro\gplbacktext{\csname LTb\endcsname \put(660,1431){\makebox(0,0)[r]{\strut{}\Large $\MLModel_1$}}\put(660,2158){\makebox(0,0)[r]{\strut{}\Large $\MLModel_2$}}\put(660,2885){\makebox(0,0)[r]{\strut{}\Large $\MLModel_4$}}\put(660,3612){\makebox(0,0)[r]{\strut{}\Large $\MLModel_5$}}\put(792,484){\makebox(0,0){\strut{}$0$}}\put(1971,484){\makebox(0,0){\strut{}$10000$}}\put(3149,484){\makebox(0,0){\strut{}$20000$}}\put(4328,484){\makebox(0,0){\strut{}$30000$}}\put(5507,484){\makebox(0,0){\strut{}$40000$}}\put(6685,484){\makebox(0,0){\strut{}$50000$}}}\gplgaddtomacro\gplfronttext{\csname LTb\endcsname \put(-77,2521){\rotatebox{-270.00}{\makebox(0,0){\strut{}\Large Models}}}\put(3797,154){\makebox(0,0){\strut{}\Large Execution traces \executionTrace}}\put(3797,4709){\makebox(0,0){\strut{}\Large \mem{5}}}}\gplbacktext
    \put(0,0){\includegraphics[width={360.00bp},height={252.00bp}]{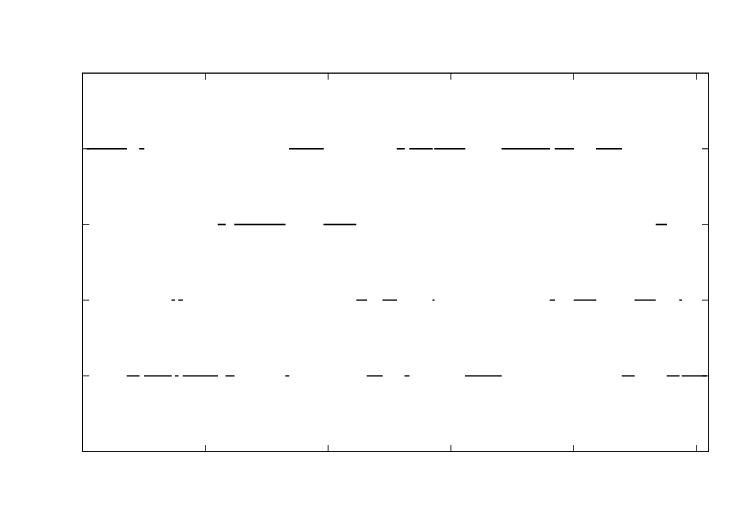}}\gplfronttext
  \end{picture}\endgroup
 				\end{adjustbox} \\
				\footnotesize (a) & \footnotesize (b) \\  
				\begin{adjustbox}{max totalsize={.95\columnwidth}{\textheight},center}
					\begingroup
  \makeatletter
  \providecommand\color[2][]{\GenericError{(gnuplot) \space\space\space\@spaces}{Package color not loaded in conjunction with
      terminal option `colourtext'}{See the gnuplot documentation for explanation.}{Either use 'blacktext' in gnuplot or load the package
      color.sty in LaTeX.}\renewcommand\color[2][]{}}\providecommand\includegraphics[2][]{\GenericError{(gnuplot) \space\space\space\@spaces}{Package graphicx or graphics not loaded}{See the gnuplot documentation for explanation.}{The gnuplot epslatex terminal needs graphicx.sty or graphics.sty.}\renewcommand\includegraphics[2][]{}}\providecommand\rotatebox[2]{#2}\@ifundefined{ifGPcolor}{\newif\ifGPcolor
    \GPcolorfalse
  }{}\@ifundefined{ifGPblacktext}{\newif\ifGPblacktext
    \GPblacktexttrue
  }{}\let\gplgaddtomacro\g@addto@macro
\gdef\gplbacktext{}\gdef\gplfronttext{}\makeatother
  \ifGPblacktext
\def\colorrgb#1{}\def\colorgray#1{}\else
\ifGPcolor
      \def\colorrgb#1{\color[rgb]{#1}}\def\colorgray#1{\color[gray]{#1}}\expandafter\def\csname LTw\endcsname{\color{white}}\expandafter\def\csname LTb\endcsname{\color{black}}\expandafter\def\csname LTa\endcsname{\color{black}}\expandafter\def\csname LT0\endcsname{\color[rgb]{1,0,0}}\expandafter\def\csname LT1\endcsname{\color[rgb]{0,1,0}}\expandafter\def\csname LT2\endcsname{\color[rgb]{0,0,1}}\expandafter\def\csname LT3\endcsname{\color[rgb]{1,0,1}}\expandafter\def\csname LT4\endcsname{\color[rgb]{0,1,1}}\expandafter\def\csname LT5\endcsname{\color[rgb]{1,1,0}}\expandafter\def\csname LT6\endcsname{\color[rgb]{0,0,0}}\expandafter\def\csname LT7\endcsname{\color[rgb]{1,0.3,0}}\expandafter\def\csname LT8\endcsname{\color[rgb]{0.5,0.5,0.5}}\else
\def\colorrgb#1{\color{black}}\def\colorgray#1{\color[gray]{#1}}\expandafter\def\csname LTw\endcsname{\color{white}}\expandafter\def\csname LTb\endcsname{\color{black}}\expandafter\def\csname LTa\endcsname{\color{black}}\expandafter\def\csname LT0\endcsname{\color{black}}\expandafter\def\csname LT1\endcsname{\color{black}}\expandafter\def\csname LT2\endcsname{\color{black}}\expandafter\def\csname LT3\endcsname{\color{black}}\expandafter\def\csname LT4\endcsname{\color{black}}\expandafter\def\csname LT5\endcsname{\color{black}}\expandafter\def\csname LT6\endcsname{\color{black}}\expandafter\def\csname LT7\endcsname{\color{black}}\expandafter\def\csname LT8\endcsname{\color{black}}\fi
  \fi
    \setlength{\unitlength}{0.0500bp}\ifx\gptboxheight\undefined \newlength{\gptboxheight}\newlength{\gptboxwidth}\newsavebox{\gptboxtext}\fi \setlength{\fboxrule}{0.5pt}\setlength{\fboxsep}{1pt}\definecolor{tbcol}{rgb}{1,1,1}\begin{picture}(7200.00,5040.00)\gplgaddtomacro\gplbacktext{\csname LTb\endcsname \put(660,1431){\makebox(0,0)[r]{\strut{}\Large $\MLModel_1$}}\put(660,2158){\makebox(0,0)[r]{\strut{}\Large $\MLModel_2$}}\put(660,2885){\makebox(0,0)[r]{\strut{}\Large $\MLModel_4$}}\put(660,3612){\makebox(0,0)[r]{\strut{}\Large $\MLModel_5$}}\put(792,484){\makebox(0,0){\strut{}$0$}}\put(1971,484){\makebox(0,0){\strut{}$10000$}}\put(3149,484){\makebox(0,0){\strut{}$20000$}}\put(4328,484){\makebox(0,0){\strut{}$30000$}}\put(5507,484){\makebox(0,0){\strut{}$40000$}}\put(6685,484){\makebox(0,0){\strut{}$50000$}}}\gplgaddtomacro\gplfronttext{\csname LTb\endcsname \put(-77,2521){\rotatebox{-270.00}{\makebox(0,0){\strut{}\Large Models}}}\put(3797,154){\makebox(0,0){\strut{}\Large Execution traces \executionTrace}}\put(3797,4709){\makebox(0,0){\strut{}\Large \mem{10}}}}\gplbacktext
    \put(0,0){\includegraphics[width={360.00bp},height={252.00bp}]{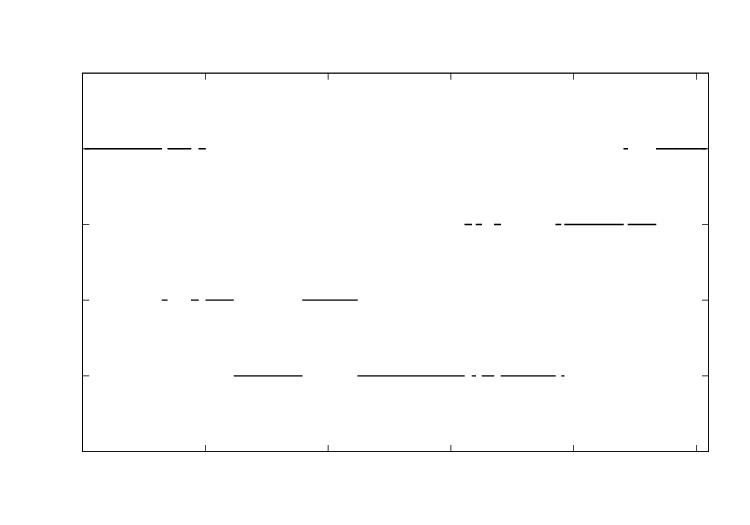}}\gplfronttext
  \end{picture}\endgroup
 				\end{adjustbox} &
				\begin{adjustbox}{max totalsize={.95\columnwidth}{\textheight},center}
					\begingroup
  \makeatletter
  \providecommand\color[2][]{\GenericError{(gnuplot) \space\space\space\@spaces}{Package color not loaded in conjunction with
      terminal option `colourtext'}{See the gnuplot documentation for explanation.}{Either use 'blacktext' in gnuplot or load the package
      color.sty in LaTeX.}\renewcommand\color[2][]{}}\providecommand\includegraphics[2][]{\GenericError{(gnuplot) \space\space\space\@spaces}{Package graphicx or graphics not loaded}{See the gnuplot documentation for explanation.}{The gnuplot epslatex terminal needs graphicx.sty or graphics.sty.}\renewcommand\includegraphics[2][]{}}\providecommand\rotatebox[2]{#2}\@ifundefined{ifGPcolor}{\newif\ifGPcolor
    \GPcolorfalse
  }{}\@ifundefined{ifGPblacktext}{\newif\ifGPblacktext
    \GPblacktexttrue
  }{}\let\gplgaddtomacro\g@addto@macro
\gdef\gplbacktext{}\gdef\gplfronttext{}\makeatother
  \ifGPblacktext
\def\colorrgb#1{}\def\colorgray#1{}\else
\ifGPcolor
      \def\colorrgb#1{\color[rgb]{#1}}\def\colorgray#1{\color[gray]{#1}}\expandafter\def\csname LTw\endcsname{\color{white}}\expandafter\def\csname LTb\endcsname{\color{black}}\expandafter\def\csname LTa\endcsname{\color{black}}\expandafter\def\csname LT0\endcsname{\color[rgb]{1,0,0}}\expandafter\def\csname LT1\endcsname{\color[rgb]{0,1,0}}\expandafter\def\csname LT2\endcsname{\color[rgb]{0,0,1}}\expandafter\def\csname LT3\endcsname{\color[rgb]{1,0,1}}\expandafter\def\csname LT4\endcsname{\color[rgb]{0,1,1}}\expandafter\def\csname LT5\endcsname{\color[rgb]{1,1,0}}\expandafter\def\csname LT6\endcsname{\color[rgb]{0,0,0}}\expandafter\def\csname LT7\endcsname{\color[rgb]{1,0.3,0}}\expandafter\def\csname LT8\endcsname{\color[rgb]{0.5,0.5,0.5}}\else
\def\colorrgb#1{\color{black}}\def\colorgray#1{\color[gray]{#1}}\expandafter\def\csname LTw\endcsname{\color{white}}\expandafter\def\csname LTb\endcsname{\color{black}}\expandafter\def\csname LTa\endcsname{\color{black}}\expandafter\def\csname LT0\endcsname{\color{black}}\expandafter\def\csname LT1\endcsname{\color{black}}\expandafter\def\csname LT2\endcsname{\color{black}}\expandafter\def\csname LT3\endcsname{\color{black}}\expandafter\def\csname LT4\endcsname{\color{black}}\expandafter\def\csname LT5\endcsname{\color{black}}\expandafter\def\csname LT6\endcsname{\color{black}}\expandafter\def\csname LT7\endcsname{\color{black}}\expandafter\def\csname LT8\endcsname{\color{black}}\fi
  \fi
    \setlength{\unitlength}{0.0500bp}\ifx\gptboxheight\undefined \newlength{\gptboxheight}\newlength{\gptboxwidth}\newsavebox{\gptboxtext}\fi \setlength{\fboxrule}{0.5pt}\setlength{\fboxsep}{1pt}\definecolor{tbcol}{rgb}{1,1,1}\begin{picture}(7200.00,5040.00)\gplgaddtomacro\gplbacktext{\csname LTb\endcsname \put(660,1431){\makebox(0,0)[r]{\strut{}\Large $\MLModel_1$}}\put(660,2158){\makebox(0,0)[r]{\strut{}\Large $\MLModel_2$}}\put(660,2885){\makebox(0,0)[r]{\strut{}\Large $\MLModel_4$}}\put(660,3612){\makebox(0,0)[r]{\strut{}\Large $\MLModel_5$}}\put(792,484){\makebox(0,0){\strut{}$0$}}\put(1971,484){\makebox(0,0){\strut{}$10000$}}\put(3149,484){\makebox(0,0){\strut{}$20000$}}\put(4328,484){\makebox(0,0){\strut{}$30000$}}\put(5507,484){\makebox(0,0){\strut{}$40000$}}\put(6685,484){\makebox(0,0){\strut{}$50000$}}}\gplgaddtomacro\gplfronttext{\csname LTb\endcsname \put(-77,2521){\rotatebox{-270.00}{\makebox(0,0){\strut{}\Large Models}}}\put(3797,154){\makebox(0,0){\strut{}\Large Execution traces \executionTrace}}\put(3797,4709){\makebox(0,0){\strut{}\mem{25}}}}\gplbacktext
    \put(0,0){\includegraphics[width={360.00bp},height={252.00bp}]{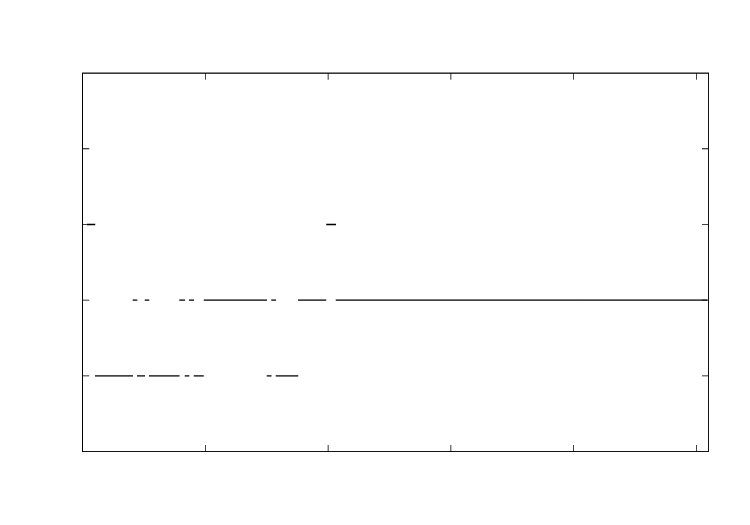}}\gplfronttext
  \end{picture}\endgroup
 				\end{adjustbox} \\

				\footnotesize (c) & \footnotesize (d) \\ 
			\end{tabular}\\
		\end{tabular}
	}
  \caption{The selected model for each execution trace \executionTrace\ of experiment $\experiment{1}$ with different memory sizes \mem{}.\label{fig:Win}}
\end{figure}

Let us now consider the model selected according to the \dynamicMABVariableWindowMemory\ ranking. Figure~\ref{fig:Win} shows the selected model for each set of execution traces in experiment $\experiment{1}$, considering different memory sizes. We note that extemporaneous changes on the selected model are frequent without memory (\mem{0}), less frequent with \mem{5}, where clusters of continuously selected models emerge, and highly infrequent with \mem{10}. Figure~\ref{fig:Win}(d) shows a stable selection of model $\MLModel_2$, while models $\MLModel_4$ and $\MLModel_5$ are often not selected preferring $\MLModel_1$ instead. Considering the entire ranking, model $\MLModel_1$ is ranked at the second position with \mem{25}, while $\MLModel_4$ at the third position. 

 In general, we observe that the number of changes across the experiments, in terms of selected models, decreases as the memory increases. On average, across all experiments, it decreases by $41.18\%$ when memory increases from \mem{5} to \mem{10} (from $34$ changes on average with \mem{5} to $20$ changes on average with \mem{10}); it decreases by $20\%$ when memory increases from \mem{10} to \mem{25} (from $20$ changes on average with \mem{10} to $16$ changes on average with \mem{25}).

\begin{figure}[!t]
	\centering
\begin{adjustbox}{max totalsize={.95\columnwidth}{\textheight},center}
		\begingroup
  \makeatletter
  \providecommand\color[2][]{\GenericError{(gnuplot) \space\space\space\@spaces}{Package color not loaded in conjunction with
      terminal option `colourtext'}{See the gnuplot documentation for explanation.}{Either use 'blacktext' in gnuplot or load the package
      color.sty in LaTeX.}\renewcommand\color[2][]{}}\providecommand\includegraphics[2][]{\GenericError{(gnuplot) \space\space\space\@spaces}{Package graphicx or graphics not loaded}{See the gnuplot documentation for explanation.}{The gnuplot epslatex terminal needs graphicx.sty or graphics.sty.}\renewcommand\includegraphics[2][]{}}\providecommand\rotatebox[2]{#2}\@ifundefined{ifGPcolor}{\newif\ifGPcolor
    \GPcolorfalse
  }{}\@ifundefined{ifGPblacktext}{\newif\ifGPblacktext
    \GPblacktexttrue
  }{}\let\gplgaddtomacro\g@addto@macro
\gdef\gplbacktext{}\gdef\gplfronttext{}\makeatother
  \ifGPblacktext
\def\colorrgb#1{}\def\colorgray#1{}\else
\ifGPcolor
      \def\colorrgb#1{\color[rgb]{#1}}\def\colorgray#1{\color[gray]{#1}}\expandafter\def\csname LTw\endcsname{\color{white}}\expandafter\def\csname LTb\endcsname{\color{black}}\expandafter\def\csname LTa\endcsname{\color{black}}\expandafter\def\csname LT0\endcsname{\color[rgb]{1,0,0}}\expandafter\def\csname LT1\endcsname{\color[rgb]{0,1,0}}\expandafter\def\csname LT2\endcsname{\color[rgb]{0,0,1}}\expandafter\def\csname LT3\endcsname{\color[rgb]{1,0,1}}\expandafter\def\csname LT4\endcsname{\color[rgb]{0,1,1}}\expandafter\def\csname LT5\endcsname{\color[rgb]{1,1,0}}\expandafter\def\csname LT6\endcsname{\color[rgb]{0,0,0}}\expandafter\def\csname LT7\endcsname{\color[rgb]{1,0.3,0}}\expandafter\def\csname LT8\endcsname{\color[rgb]{0.5,0.5,0.5}}\else
\def\colorrgb#1{\color{black}}\def\colorgray#1{\color[gray]{#1}}\expandafter\def\csname LTw\endcsname{\color{white}}\expandafter\def\csname LTb\endcsname{\color{black}}\expandafter\def\csname LTa\endcsname{\color{black}}\expandafter\def\csname LT0\endcsname{\color{black}}\expandafter\def\csname LT1\endcsname{\color{black}}\expandafter\def\csname LT2\endcsname{\color{black}}\expandafter\def\csname LT3\endcsname{\color{black}}\expandafter\def\csname LT4\endcsname{\color{black}}\expandafter\def\csname LT5\endcsname{\color{black}}\expandafter\def\csname LT6\endcsname{\color{black}}\expandafter\def\csname LT7\endcsname{\color{black}}\expandafter\def\csname LT8\endcsname{\color{black}}\fi
  \fi
    \setlength{\unitlength}{0.0500bp}\ifx\gptboxheight\undefined \newlength{\gptboxheight}\newlength{\gptboxwidth}\newsavebox{\gptboxtext}\fi \setlength{\fboxrule}{0.5pt}\setlength{\fboxsep}{1pt}\definecolor{tbcol}{rgb}{1,1,1}\begin{picture}(7200.00,5040.00)\gplgaddtomacro\gplbacktext{\csname LTb\endcsname \put(814,858){\makebox(0,0)[r]{\strut{}$0$}}\csname LTb\endcsname \put(814,1482){\makebox(0,0)[r]{\strut{}$20$}}\csname LTb\endcsname \put(814,2105){\makebox(0,0)[r]{\strut{}$40$}}\csname LTb\endcsname \put(814,2729){\makebox(0,0)[r]{\strut{}$60$}}\csname LTb\endcsname \put(814,3352){\makebox(0,0)[r]{\strut{}$80$}}\csname LTb\endcsname \put(814,3976){\makebox(0,0)[r]{\strut{}$100$}}\put(1089,726){\rotatebox{-270.00}{\makebox(0,0)[r]{\strut{}$\MLModel_1$}}}\put(1232,726){\rotatebox{-270.00}{\makebox(0,0)[r]{\strut{}$\MLModel_2$}}}\put(1375,726){\rotatebox{-270.00}{\makebox(0,0)[r]{\strut{}$\MLModel_4$}}}\put(1517,726){\rotatebox{-270.00}{\makebox(0,0)[r]{\strut{}$\MLModel_5$}}}\put(1660,726){\rotatebox{-270.00}{\makebox(0,0)[r]{\strut{}$\MLModel_1$}}}\put(1803,726){\rotatebox{-270.00}{\makebox(0,0)[r]{\strut{}$\MLModel_2$}}}\put(1946,726){\rotatebox{-270.00}{\makebox(0,0)[r]{\strut{}$\MLModel_4$}}}\put(2089,726){\rotatebox{-270.00}{\makebox(0,0)[r]{\strut{}$\MLModel_5$}}}\put(2232,726){\rotatebox{-270.00}{\makebox(0,0)[r]{\strut{}$\MLModel_1$}}}\put(2375,726){\rotatebox{-270.00}{\makebox(0,0)[r]{\strut{}$\MLModel_2$}}}\put(2517,726){\rotatebox{-270.00}{\makebox(0,0)[r]{\strut{}$\MLModel_4$}}}\put(2660,726){\rotatebox{-270.00}{\makebox(0,0)[r]{\strut{}$\MLModel_5$}}}\put(2803,726){\rotatebox{-270.00}{\makebox(0,0)[r]{\strut{}$\MLModel_1$}}}\put(2946,726){\rotatebox{-270.00}{\makebox(0,0)[r]{\strut{}$\MLModel_2$}}}\put(3089,726){\rotatebox{-270.00}{\makebox(0,0)[r]{\strut{}$\MLModel_4$}}}\put(3232,726){\rotatebox{-270.00}{\makebox(0,0)[r]{\strut{}$\MLModel_5$}}}\put(3375,726){\rotatebox{-270.00}{\makebox(0,0)[r]{\strut{}$\MLModel_1$}}}\put(3517,726){\rotatebox{-270.00}{\makebox(0,0)[r]{\strut{}$\MLModel_2$}}}\put(3660,726){\rotatebox{-270.00}{\makebox(0,0)[r]{\strut{}$\MLModel_4$}}}\put(3803,726){\rotatebox{-270.00}{\makebox(0,0)[r]{\strut{}$\MLModel_5$}}}\put(3946,726){\rotatebox{-270.00}{\makebox(0,0)[r]{\strut{}$\MLModel_1$}}}\put(4089,726){\rotatebox{-270.00}{\makebox(0,0)[r]{\strut{}$\MLModel_2$}}}\put(4232,726){\rotatebox{-270.00}{\makebox(0,0)[r]{\strut{}$\MLModel_4$}}}\put(4374,726){\rotatebox{-270.00}{\makebox(0,0)[r]{\strut{}$\MLModel_5$}}}\put(4517,726){\rotatebox{-270.00}{\makebox(0,0)[r]{\strut{}$\MLModel_1$}}}\put(4660,726){\rotatebox{-270.00}{\makebox(0,0)[r]{\strut{}$\MLModel_2$}}}\put(4803,726){\rotatebox{-270.00}{\makebox(0,0)[r]{\strut{}$\MLModel_4$}}}\put(4946,726){\rotatebox{-270.00}{\makebox(0,0)[r]{\strut{}$\MLModel_5$}}}\put(5089,726){\rotatebox{-270.00}{\makebox(0,0)[r]{\strut{}$\MLModel_1$}}}\put(5232,726){\rotatebox{-270.00}{\makebox(0,0)[r]{\strut{}$\MLModel_2$}}}\put(5374,726){\rotatebox{-270.00}{\makebox(0,0)[r]{\strut{}$\MLModel_4$}}}\put(5517,726){\rotatebox{-270.00}{\makebox(0,0)[r]{\strut{}$\MLModel_5$}}}\put(5660,726){\rotatebox{-270.00}{\makebox(0,0)[r]{\strut{}$\MLModel_1$}}}\put(5803,726){\rotatebox{-270.00}{\makebox(0,0)[r]{\strut{}$\MLModel_2$}}}\put(5946,726){\rotatebox{-270.00}{\makebox(0,0)[r]{\strut{}$\MLModel_4$}}}\put(6089,726){\rotatebox{-270.00}{\makebox(0,0)[r]{\strut{}$\MLModel_5$}}}\put(6232,726){\rotatebox{-270.00}{\makebox(0,0)[r]{\strut{}$\MLModel_1$}}}\put(6374,726){\rotatebox{-270.00}{\makebox(0,0)[r]{\strut{}$\MLModel_2$}}}\put(6517,726){\rotatebox{-270.00}{\makebox(0,0)[r]{\strut{}$\MLModel_4$}}}\put(6660,726){\rotatebox{-270.00}{\makebox(0,0)[r]{\strut{}$\MLModel_5$}}}}\gplgaddtomacro\gplfronttext{\csname LTb\endcsname \put(1768,4867){\makebox(0,0)[r]{\strut{}1st}}\csname LTb\endcsname \put(3019,4867){\makebox(0,0)[r]{\strut{}2nd}}\csname LTb\endcsname \put(4270,4867){\makebox(0,0)[r]{\strut{}3rd}}\csname LTb\endcsname \put(5521,4867){\makebox(0,0)[r]{\strut{}4th}}\csname LTb\endcsname \put(209,2728){\rotatebox{-270.00}{\makebox(0,0){\strut{}Ranking (\%)}}}\put(3874,154){\makebox(0,0){\strut{}Experiments}}\colorrgb{0.58,0.00,0.83}\put(1103,4287){\makebox(0,0)[l]{\strut{}\experiment{1}}}\put(1675,4287){\makebox(0,0)[l]{\strut{}\experiment{2}}}\put(2246,4287){\makebox(0,0)[l]{\strut{}\experiment{3}}}\put(2817,4287){\makebox(0,0)[l]{\strut{}\experiment{4}}}\put(3389,4287){\makebox(0,0)[l]{\strut{}\experiment{5}}}\put(3960,4287){\makebox(0,0)[l]{\strut{}\experiment{6}}}\put(4532,4287){\makebox(0,0)[l]{\strut{}\experiment{7}}}\put(5103,4287){\makebox(0,0)[l]{\strut{}\experiment{8}}}\put(5674,4287){\makebox(0,0)[l]{\strut{}\experiment{9}}}\put(6246,4287){\makebox(0,0)[l]{\strut{}\experiment{10}}}}\gplbacktext
    \put(0,0){\includegraphics[width={360.00bp},height={252.00bp}]{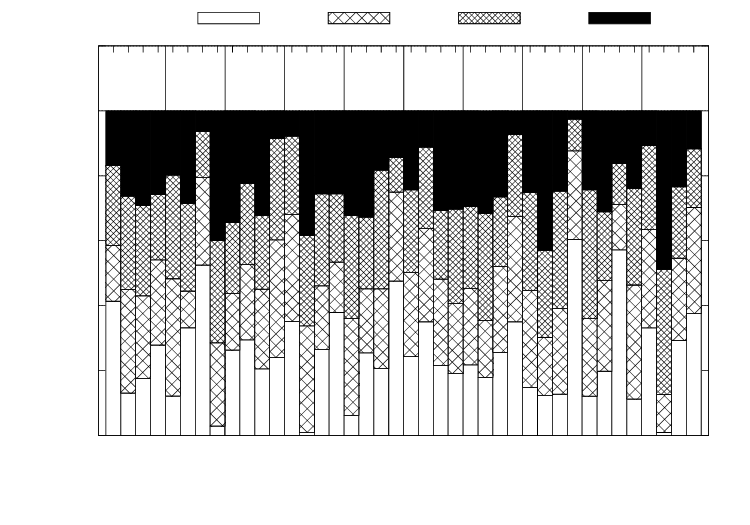}}\gplfronttext
  \end{picture}\endgroup
 	\end{adjustbox}
	\caption{Stacked histograms showing the ranking of the models in each of the 10 experiments with \mem{10}.}
	\label{fig:hist}
\end{figure}

Figure~\ref{fig:hist} shows an aggregated ranking for all the experiments with \mem{10}. It shows the percentage of times a model has been ranked into a specific position for all the experiments. 
We note that $\experiment{1}$ was one of the most balanced experiments in terms of ranking, having at least three models ($\MLModel_1$, $\MLModel_2$, and $\MLModel_5$) with a similar percentage of first and second positions in the ranking. In \experiment{2} and \experiment{8}, $\MLModel_4$ and $\MLModel_5$ were ranked at first or second position $\approx$$80\%$ of the times. 
More specifically, $\MLModel_1$ and $\MLModel_5$ are ranked as the first model in the ranking $41.39$\% and $27.87$\% of times, respectively (Figure~\ref{fig:Win}(c)). When $\MLModel_5$ is not ranked first, it is ranked second $26.36\%$ of times, while $\MLModel_1$ $17.21\%$.

Considering all experiments and memory sizes, when compared with \mem{5}, we note an average decrease of ranking changes in the first position of $34.66$\% with \mem{10} and of $62.43$\% with \mem{25}. This is also clear from Figure~\ref{fig:Win}(d), where $\MLModel_2$ was ranked in the first position most of the times.

\subsubsection{Quality Evaluation}\label{subsubsec:experiments-substitution-quality}

We evaluated the quality of ranking-based substitution and assurance-based substitution varying the memory.  The ranking retrieved according to \dynamicMABVariableWindow\ is used as baseline.

Let \rankingFunc\ denote the function that returns as output the (current) position in the \dynamicMABVariableWindow\-/based ranking of the (current) top\-/ranked model in the \dynamicMABVariableWindowMemory\-/based ranking; this position is a number $\in$$[1,$ \ldots$,$ $k]$ with $k$$=$$\vert\candidateList\vert$. The \emph{residual error} $\err$ measures the difference between the ranking obtained with \dynamicMABVariableWindowMemory\ and \dynamicMABVariableWindow, and is defined as:

\begin{equation} 
	\err=\penaltyFunc\left(\frac{\rankingFunc{}-1}{k-1}\right), \label{eq:residual}
\end{equation}

where \penaltyFunc{} is the \emph{residual penalty function}. We note that in case the top\-/ranked model according to \dynamicMABVariableWindowMemory\ is top\-/ranked also according to \dynamicMABVariableWindow, the residual error is \err$=$$\penaltyFunc(0)$; in case it is ranked last according to \dynamicMABVariableWindow, the residual error is \err$=$$\penaltyFunc(1)$. Residual penalty function \penaltyFunc\ is defined as a sigmoid function as follows:

\begin{equation} 
  \penaltyFunc(x) = \frac{1}{1+e^{-c_1(x-c_2)}}, \label{eq:penalty}
\end{equation}

where $c_2$ control the $x$ of the sigmoid inflection point and $c_1$ the slope.
The residual error measures the difference in terms of ranking between the different settings. While it is not an indicator of the absolute quality, we assume this measure as a valid indicator of the relative quality between the different settings of our solution.

\vspace{1em}

\begin{figure}[!t]
    \centering
	\begin{adjustbox}{max totalsize={.95\columnwidth}{\textheight},center}
		\begingroup
  \makeatletter
  \providecommand\color[2][]{\GenericError{(gnuplot) \space\space\space\@spaces}{Package color not loaded in conjunction with
      terminal option `colourtext'}{See the gnuplot documentation for explanation.}{Either use 'blacktext' in gnuplot or load the package
      color.sty in LaTeX.}\renewcommand\color[2][]{}}\providecommand\includegraphics[2][]{\GenericError{(gnuplot) \space\space\space\@spaces}{Package graphicx or graphics not loaded}{See the gnuplot documentation for explanation.}{The gnuplot epslatex terminal needs graphicx.sty or graphics.sty.}\renewcommand\includegraphics[2][]{}}\providecommand\rotatebox[2]{#2}\@ifundefined{ifGPcolor}{\newif\ifGPcolor
    \GPcolorfalse
  }{}\@ifundefined{ifGPblacktext}{\newif\ifGPblacktext
    \GPblacktexttrue
  }{}\let\gplgaddtomacro\g@addto@macro
\gdef\gplbacktext{}\gdef\gplfronttext{}\makeatother
  \ifGPblacktext
\def\colorrgb#1{}\def\colorgray#1{}\else
\ifGPcolor
      \def\colorrgb#1{\color[rgb]{#1}}\def\colorgray#1{\color[gray]{#1}}\expandafter\def\csname LTw\endcsname{\color{white}}\expandafter\def\csname LTb\endcsname{\color{black}}\expandafter\def\csname LTa\endcsname{\color{black}}\expandafter\def\csname LT0\endcsname{\color[rgb]{1,0,0}}\expandafter\def\csname LT1\endcsname{\color[rgb]{0,1,0}}\expandafter\def\csname LT2\endcsname{\color[rgb]{0,0,1}}\expandafter\def\csname LT3\endcsname{\color[rgb]{1,0,1}}\expandafter\def\csname LT4\endcsname{\color[rgb]{0,1,1}}\expandafter\def\csname LT5\endcsname{\color[rgb]{1,1,0}}\expandafter\def\csname LT6\endcsname{\color[rgb]{0,0,0}}\expandafter\def\csname LT7\endcsname{\color[rgb]{1,0.3,0}}\expandafter\def\csname LT8\endcsname{\color[rgb]{0.5,0.5,0.5}}\else
\def\colorrgb#1{\color{black}}\def\colorgray#1{\color[gray]{#1}}\expandafter\def\csname LTw\endcsname{\color{white}}\expandafter\def\csname LTb\endcsname{\color{black}}\expandafter\def\csname LTa\endcsname{\color{black}}\expandafter\def\csname LT0\endcsname{\color{black}}\expandafter\def\csname LT1\endcsname{\color{black}}\expandafter\def\csname LT2\endcsname{\color{black}}\expandafter\def\csname LT3\endcsname{\color{black}}\expandafter\def\csname LT4\endcsname{\color{black}}\expandafter\def\csname LT5\endcsname{\color{black}}\expandafter\def\csname LT6\endcsname{\color{black}}\expandafter\def\csname LT7\endcsname{\color{black}}\expandafter\def\csname LT8\endcsname{\color{black}}\fi
  \fi
    \setlength{\unitlength}{0.0500bp}\ifx\gptboxheight\undefined \newlength{\gptboxheight}\newlength{\gptboxwidth}\newsavebox{\gptboxtext}\fi \setlength{\fboxrule}{0.5pt}\setlength{\fboxsep}{1pt}\definecolor{tbcol}{rgb}{1,1,1}\begin{picture}(7200.00,5040.00)\gplgaddtomacro\gplbacktext{\csname LTb\endcsname \put(814,704){\makebox(0,0)[r]{\strut{}$0$}}\put(814,1390){\makebox(0,0)[r]{\strut{}$50$}}\put(814,2076){\makebox(0,0)[r]{\strut{}$100$}}\put(814,2762){\makebox(0,0)[r]{\strut{}$150$}}\put(814,3447){\makebox(0,0)[r]{\strut{}$200$}}\put(814,4133){\makebox(0,0)[r]{\strut{}$250$}}\put(814,4819){\makebox(0,0)[r]{\strut{}$300$}}\put(946,484){\makebox(0,0){\strut{}$0$}}\put(2094,484){\makebox(0,0){\strut{}$10000$}}\put(3243,484){\makebox(0,0){\strut{}$20000$}}\put(4391,484){\makebox(0,0){\strut{}$30000$}}\put(5540,484){\makebox(0,0){\strut{}$40000$}}\put(6688,484){\makebox(0,0){\strut{}$50000$}}}\gplgaddtomacro\gplfronttext{\csname LTb\endcsname \put(5816,4591){\makebox(0,0)[r]{\strut{}\mem{5} }}\put(5816,4261){\makebox(0,0)[r]{\strut{}\mem{10}}}\put(5816,3931){\makebox(0,0)[r]{\strut{}\mem{25}}}\put(209,2761){\rotatebox{-270.00}{\makebox(0,0){\strut{}Cumulative error \cumulativeError}}}\put(3874,154){\makebox(0,0){\strut{}Execution trace \executionTrace}}}\gplbacktext
    \put(0,0){\includegraphics[width={360.00bp},height={252.00bp}]{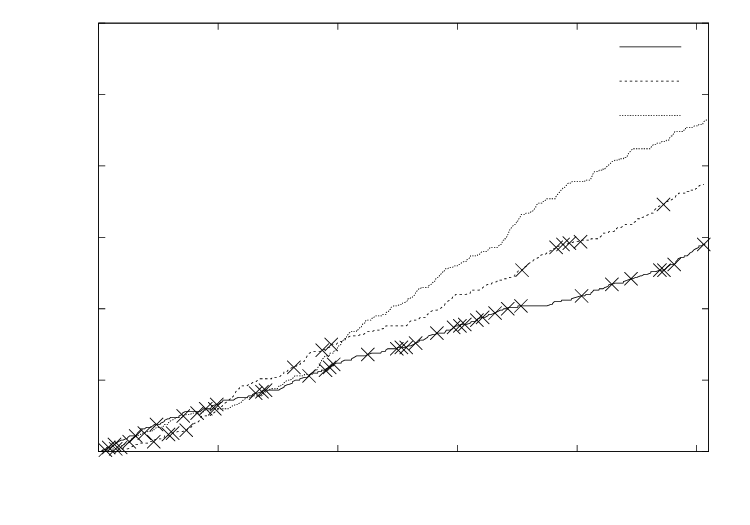}}\gplfronttext
  \end{picture}\endgroup
 	\end{adjustbox}
\caption{Cumulative residual error $\hat{\err}_k$ between DMVW ($\mem{0}$) and the \dynamicMABVariableWindowMemory\ with different memory sizes $\mem{i}$ and number of traces $k$ for experiment $\experiment{1}$. Model substitutions are marked with ``X''.}
	\label{fig:cumulative}
\end{figure}

\noindent \textbf{Ranking-based substitution}: 
 Figure~\ref{fig:cumulative} shows the \emph{cumulative residual error} \cumulativeError$=$$\sum_{\instant} \err$ (i.e., the sum of the error retrieved in each window and execution trace) for $\experiment{1}$ with different memory sizes.
It also shows, marked with ``$\times$'', the execution traces where model substitutions occurred due to changes at the top of the ranking. We note that in this experiment the bigger the memory, the bigger the cumulative residual error. This effect is compensated by fewer model substitutions as also demonstrated in Section~\ref{subsubsec:experiments-substitution-memory}. 
We also note that, depending on the application domain, the memory settings can be dynamic. For instance, in scenarios where fast reaction to changes is more important than stability of the selected model, the memory can be lowered; it can be increased in scenarios where stability is important to counteract fluctuations.

Considering all experiments and memory sizes, we note an average cumulative residual error of $163.67$ with \mem{5} and of $243.63$ with \mem{25}, corresponding to an average increase of $48.85\%$. As depicted in Figure~\ref{fig:hist}, our experiments revealed a frequent variability among the best candidate models. Therefore, the most suitable approach in terms of residual error was the one having lower memory.

\vspace{1em}

\noindent \textbf{Assurance-based substitution}: 
Using the memory settings in Section~\ref{subsubsec:experiments-substitution-memory}, we first evaluated the impact of the degradation threshold \thresholdAssurance, varying its value in $\thresholdAssurance_5$$=$$0.05$, $\thresholdAssurance_{10}$$=$$0.10$, and $\thresholdAssurance_{25}$$=$$0.25$.

\begin{figure}[!t]
\begin{adjustbox}{max totalsize={.95\columnwidth}{\textheight},center}
		\begingroup
  \makeatletter
  \providecommand\color[2][]{\GenericError{(gnuplot) \space\space\space\@spaces}{Package color not loaded in conjunction with
      terminal option `colourtext'}{See the gnuplot documentation for explanation.}{Either use 'blacktext' in gnuplot or load the package
      color.sty in LaTeX.}\renewcommand\color[2][]{}}\providecommand\includegraphics[2][]{\GenericError{(gnuplot) \space\space\space\@spaces}{Package graphicx or graphics not loaded}{See the gnuplot documentation for explanation.}{The gnuplot epslatex terminal needs graphicx.sty or graphics.sty.}\renewcommand\includegraphics[2][]{}}\providecommand\rotatebox[2]{#2}\@ifundefined{ifGPcolor}{\newif\ifGPcolor
    \GPcolorfalse
  }{}\@ifundefined{ifGPblacktext}{\newif\ifGPblacktext
    \GPblacktexttrue
  }{}\let\gplgaddtomacro\g@addto@macro
\gdef\gplbacktext{}\gdef\gplfronttext{}\makeatother
  \ifGPblacktext
\def\colorrgb#1{}\def\colorgray#1{}\else
\ifGPcolor
      \def\colorrgb#1{\color[rgb]{#1}}\def\colorgray#1{\color[gray]{#1}}\expandafter\def\csname LTw\endcsname{\color{white}}\expandafter\def\csname LTb\endcsname{\color{black}}\expandafter\def\csname LTa\endcsname{\color{black}}\expandafter\def\csname LT0\endcsname{\color[rgb]{1,0,0}}\expandafter\def\csname LT1\endcsname{\color[rgb]{0,1,0}}\expandafter\def\csname LT2\endcsname{\color[rgb]{0,0,1}}\expandafter\def\csname LT3\endcsname{\color[rgb]{1,0,1}}\expandafter\def\csname LT4\endcsname{\color[rgb]{0,1,1}}\expandafter\def\csname LT5\endcsname{\color[rgb]{1,1,0}}\expandafter\def\csname LT6\endcsname{\color[rgb]{0,0,0}}\expandafter\def\csname LT7\endcsname{\color[rgb]{1,0.3,0}}\expandafter\def\csname LT8\endcsname{\color[rgb]{0.5,0.5,0.5}}\else
\def\colorrgb#1{\color{black}}\def\colorgray#1{\color[gray]{#1}}\expandafter\def\csname LTw\endcsname{\color{white}}\expandafter\def\csname LTb\endcsname{\color{black}}\expandafter\def\csname LTa\endcsname{\color{black}}\expandafter\def\csname LT0\endcsname{\color{black}}\expandafter\def\csname LT1\endcsname{\color{black}}\expandafter\def\csname LT2\endcsname{\color{black}}\expandafter\def\csname LT3\endcsname{\color{black}}\expandafter\def\csname LT4\endcsname{\color{black}}\expandafter\def\csname LT5\endcsname{\color{black}}\expandafter\def\csname LT6\endcsname{\color{black}}\expandafter\def\csname LT7\endcsname{\color{black}}\expandafter\def\csname LT8\endcsname{\color{black}}\fi
  \fi
    \setlength{\unitlength}{0.0500bp}\ifx\gptboxheight\undefined \newlength{\gptboxheight}\newlength{\gptboxwidth}\newsavebox{\gptboxtext}\fi \setlength{\fboxrule}{0.5pt}\setlength{\fboxsep}{1pt}\definecolor{tbcol}{rgb}{1,1,1}\begin{picture}(7200.00,5040.00)\gplgaddtomacro\gplbacktext{\csname LTb\endcsname \put(814,704){\makebox(0,0)[r]{\strut{}$0$}}\csname LTb\endcsname \put(814,1368){\makebox(0,0)[r]{\strut{}$50$}}\csname LTb\endcsname \put(814,2031){\makebox(0,0)[r]{\strut{}$100$}}\csname LTb\endcsname \put(814,2695){\makebox(0,0)[r]{\strut{}$150$}}\csname LTb\endcsname \put(814,3359){\makebox(0,0)[r]{\strut{}$200$}}\csname LTb\endcsname \put(814,4023){\makebox(0,0)[r]{\strut{}$250$}}\csname LTb\endcsname \put(814,4686){\makebox(0,0)[r]{\strut{}$300$}}\put(1466,484){\makebox(0,0){\strut{}$\mem{5}$}}\put(2051,484){\makebox(0,0){\strut{}$\mem{10}$}}\put(2637,484){\makebox(0,0){\strut{}$\mem{25}$}}\put(3223,484){\makebox(0,0){\strut{}$\mem{5}$}}\put(3809,484){\makebox(0,0){\strut{}$\mem{10}$}}\put(4394,484){\makebox(0,0){\strut{}$\mem{25}$}}\put(4980,484){\makebox(0,0){\strut{}$\mem{5}$}}\put(5566,484){\makebox(0,0){\strut{}$\mem{10}$}}\put(6151,484){\makebox(0,0){\strut{}$\mem{25}$}}\put(2117,154){\makebox(0,0){\strut{}$\thresholdAssurance_{5}$}}\put(3875,154){\makebox(0,0){\strut{}$\thresholdAssurance_{10}$}}\put(5632,154){\makebox(0,0){\strut{}$\thresholdAssurance_{25}$}}}\gplgaddtomacro\gplfronttext{\csname LTb\endcsname \put(5816,4646){\makebox(0,0)[r]{\strut{}total}}\csname LTb\endcsname \put(5816,4426){\makebox(0,0)[r]{\strut{}relevant}}\csname LTb\endcsname \put(5816,4206){\makebox(0,0)[r]{\strut{}success}}\csname LTb\endcsname \put(209,2761){\rotatebox{-270.00}{\makebox(0,0){\strut{}N$^o$ of early substitutions}}}\put(3874,154){\makebox(0,0){\strut{} }}}\gplbacktext
    \put(0,0){\includegraphics[width={360.00bp},height={252.00bp}]{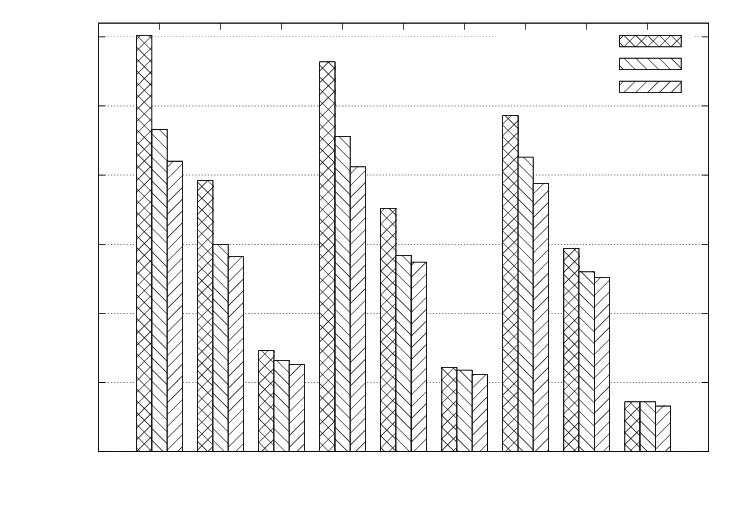}}\gplfronttext
  \end{picture}\endgroup
 	\end{adjustbox}
\caption{
Total number of triggered early substitutions (total), relevant early substitutions (relevant), and successful early substitutions (success) varying threshold \thresholdAssurance\ and memory \mem{} for \experiment{1}--\experiment{10}.}
	\label{fig:early-substitution}
\end{figure}

Figure~\ref{fig:early-substitution} shows the total number of triggered early substitutions (denoted as \emph{total}) compared with total number of substitutions that really occurred at end of the window (i.e, the correct early substitutions, denoted as \emph{relevant}), on average across all experiments and memory settings. We observe that in $89\%$ of the cases, an early substitution has been correctly triggered.  In detail, an early substitution was correctly triggered in $81\%$ of the cases when using $\thresholdAssurance_5$, increasing to $92\%$ when using $\thresholdAssurance_{25}$. These results were expected, since a higher threshold corresponds to a more severe assurance variation, and thus to a higher likelihood of the change being correct at the end of the window.

Figure~\ref{fig:early-substitution} also shows the number of successfully executed early substitutions (denoted as \emph{success}) among the \emph{relevant} early substitutions.  A \emph{successful} early substitution is a substitution where the model selected for substitution is the one evaluated by \dynamicMABVariableWindowMemory\ at the first position of the ranking at the end of the window. 
We observe that, in $93\%$ of the cases on average, assurance\-/based early substitution took the correct decision. This result also confirms the quality of the entire retrieved ranking, meaning that when a substitution was needed, the second\-/ranked model was indeed the most suitable for substitution.

We also observe that \begin{enumerate*}
	\item as the early substitution threshold \thresholdAssurance\ increases, the number of early substitutions decreases. For instance, with \mem{10}, it decreases from $196$ with $\thresholdAssurance_5$ to $147$ with $\thresholdAssurance_{25}$;
	\item the difference between the number of substitutions with $\thresholdAssurance_5$ and $\thresholdAssurance_{25}$ is lower than expected (e.g., from $301$ to $243$ with \mem{5}).
\end{enumerate*}
In other words, when a degradation occurs, it exceeds $\thresholdAssurance_{25}$ in most of the cases. Even this experiment confirms that a bigger memory corresponds to fewer early substitutions. A more stable trend of the assurance level of the selected model was also observed with bigger memory.

\begin{figure}[!t]
\begin{adjustbox}{max totalsize={.95\columnwidth}{\textheight},center}
		\begingroup
  \makeatletter
  \providecommand\color[2][]{\GenericError{(gnuplot) \space\space\space\@spaces}{Package color not loaded in conjunction with
      terminal option `colourtext'}{See the gnuplot documentation for explanation.}{Either use 'blacktext' in gnuplot or load the package
      color.sty in LaTeX.}\renewcommand\color[2][]{}}\providecommand\includegraphics[2][]{\GenericError{(gnuplot) \space\space\space\@spaces}{Package graphicx or graphics not loaded}{See the gnuplot documentation for explanation.}{The gnuplot epslatex terminal needs graphicx.sty or graphics.sty.}\renewcommand\includegraphics[2][]{}}\providecommand\rotatebox[2]{#2}\@ifundefined{ifGPcolor}{\newif\ifGPcolor
    \GPcolorfalse
  }{}\@ifundefined{ifGPblacktext}{\newif\ifGPblacktext
    \GPblacktexttrue
  }{}\let\gplgaddtomacro\g@addto@macro
\gdef\gplbacktext{}\gdef\gplfronttext{}\makeatother
  \ifGPblacktext
\def\colorrgb#1{}\def\colorgray#1{}\else
\ifGPcolor
      \def\colorrgb#1{\color[rgb]{#1}}\def\colorgray#1{\color[gray]{#1}}\expandafter\def\csname LTw\endcsname{\color{white}}\expandafter\def\csname LTb\endcsname{\color{black}}\expandafter\def\csname LTa\endcsname{\color{black}}\expandafter\def\csname LT0\endcsname{\color[rgb]{1,0,0}}\expandafter\def\csname LT1\endcsname{\color[rgb]{0,1,0}}\expandafter\def\csname LT2\endcsname{\color[rgb]{0,0,1}}\expandafter\def\csname LT3\endcsname{\color[rgb]{1,0,1}}\expandafter\def\csname LT4\endcsname{\color[rgb]{0,1,1}}\expandafter\def\csname LT5\endcsname{\color[rgb]{1,1,0}}\expandafter\def\csname LT6\endcsname{\color[rgb]{0,0,0}}\expandafter\def\csname LT7\endcsname{\color[rgb]{1,0.3,0}}\expandafter\def\csname LT8\endcsname{\color[rgb]{0.5,0.5,0.5}}\else
\def\colorrgb#1{\color{black}}\def\colorgray#1{\color[gray]{#1}}\expandafter\def\csname LTw\endcsname{\color{white}}\expandafter\def\csname LTb\endcsname{\color{black}}\expandafter\def\csname LTa\endcsname{\color{black}}\expandafter\def\csname LT0\endcsname{\color{black}}\expandafter\def\csname LT1\endcsname{\color{black}}\expandafter\def\csname LT2\endcsname{\color{black}}\expandafter\def\csname LT3\endcsname{\color{black}}\expandafter\def\csname LT4\endcsname{\color{black}}\expandafter\def\csname LT5\endcsname{\color{black}}\expandafter\def\csname LT6\endcsname{\color{black}}\expandafter\def\csname LT7\endcsname{\color{black}}\expandafter\def\csname LT8\endcsname{\color{black}}\fi
  \fi
    \setlength{\unitlength}{0.0500bp}\ifx\gptboxheight\undefined \newlength{\gptboxheight}\newlength{\gptboxwidth}\newsavebox{\gptboxtext}\fi \setlength{\fboxrule}{0.5pt}\setlength{\fboxsep}{1pt}\definecolor{tbcol}{rgb}{1,1,1}\begin{picture}(7200.00,5040.00)\gplgaddtomacro\gplbacktext{\csname LTb\endcsname \put(1342,759){\makebox(0,0)[r]{\strut{}$0$}}\csname LTb\endcsname \put(1342,1210){\makebox(0,0)[r]{\strut{}$500000$}}\csname LTb\endcsname \put(1342,1661){\makebox(0,0)[r]{\strut{}$1\times10^{6}$}}\csname LTb\endcsname \put(1342,2112){\makebox(0,0)[r]{\strut{}$1.5\times10^{6}$}}\csname LTb\endcsname \put(1342,2563){\makebox(0,0)[r]{\strut{}$2\times10^{6}$}}\csname LTb\endcsname \put(1342,3015){\makebox(0,0)[r]{\strut{}$2.5\times10^{6}$}}\csname LTb\endcsname \put(1342,3466){\makebox(0,0)[r]{\strut{}$3\times10^{6}$}}\csname LTb\endcsname \put(1342,3917){\makebox(0,0)[r]{\strut{}$3.5\times10^{6}$}}\csname LTb\endcsname \put(1342,4368){\makebox(0,0)[r]{\strut{}$4\times10^{6}$}}\csname LTb\endcsname \put(1342,4819){\makebox(0,0)[r]{\strut{}$4.5\times10^{6}$}}\put(2740,484){\makebox(0,0){\strut{}$\mem{5}$}}\put(4073,484){\makebox(0,0){\strut{}$\mem{10}$}}\put(5405,484){\makebox(0,0){\strut{}$\mem{25}$}}}\gplgaddtomacro\gplfronttext{\csname LTb\endcsname \put(5816,4646){\makebox(0,0)[r]{\strut{}$\thresholdAssurance_5$}}\csname LTb\endcsname \put(5816,4426){\makebox(0,0)[r]{\strut{}$\thresholdAssurance_{10}$}}\csname LTb\endcsname \put(5816,4206){\makebox(0,0)[r]{\strut{}$\thresholdAssurance_{25}$}}\csname LTb\endcsname \put(5816,3986){\makebox(0,0)[r]{\strut{}ranking-based}}\csname LTb\endcsname \put(143,2789){\rotatebox{-270.00}{\makebox(0,0){\strut{}\makecell{Avg time to compute an\\evaluation window (ms)}}}}\put(4138,154){\makebox(0,0){\strut{}Memory}}}\gplbacktext
    \put(0,0){\includegraphics[width={360.00bp},height={252.00bp}]{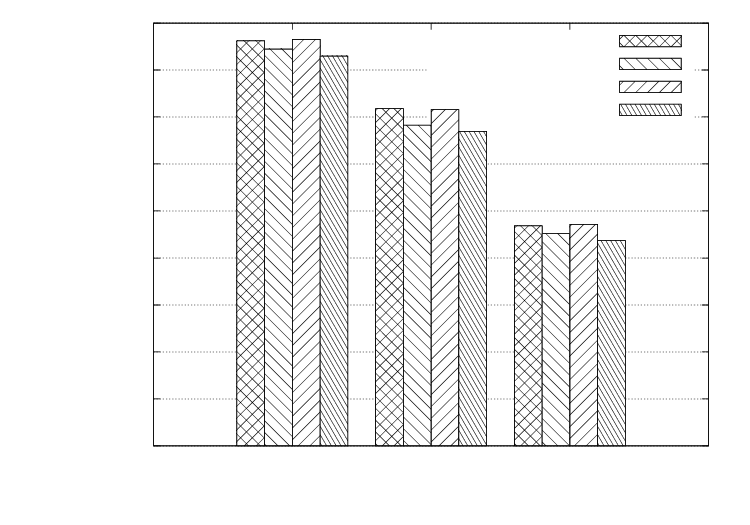}}\gplfronttext
  \end{picture}\endgroup
 	\end{adjustbox}
\caption{Performance (average windows duration) expressed in milliseconds varying threshold \thresholdAssurance\ and memory \mem{} for \experiment{1}--\experiment{12}.}
	\label{fig:perf}
\end{figure}

We then evaluated the duration of early substitutions, in terms of the number of execution traces from the moment when the early substitution is triggered to the end of the considered window.

We observe that an increase in the memory and threshold results in a decrease in the number of substitutions (see Figure~\ref{fig:early-substitution}) and their duration. On average across all experiments, the duration varies from $\approx$$290$ execution traces when using \mem{5} and $\thresholdAssurance_{5}$ to $84$ when using \mem{25} and $\thresholdAssurance_{25}$. 

 Finally, we observe that, similarly to the memory tuning, the substitution
threshold should be fine\-/tuned according to the different application domains, to adequately react to changes occurred within a given window.

\vspace{1em}

\noindent \textbf{Performance}: We compared the performance of our ranking-based and assurance-based substitutions with different memory settings and assurance thresholds on all the experiments.

Figure~\ref{fig:perf} shows both the ranking-based and the assurance-based substitution performance varying memory settings and thresholds. The results are presented as the average time to compute an evaluation window. 
We note that the ranking-based approach outperformed the assurance-based approach with an average improvement around $4.68\%$, due to the absence of assurance metric computations and corresponding comparisons. We also note that the impact of the different thresholds on performance is negligible, with $\thresholdAssurance_{10}$ showing the best performance in all conditions. Figure~\ref{fig:perf} clarifies that the dominating factor is the memory size. This is due to the fact that a bigger memory corresponds to a smaller window. In addition, it also corresponds to fewer substitutions positively impacting performance, because each substitution requires more iterations in \dynamicMABVariableWindowMemory\ to converge.

 \section{Related Work}\label{sec:related}

Our approach guarantees stable application behavior over time, by dynamically selecting the most suitable ML model according to a (set of) non\-/functional property. This issue has been studied from different angles in the literature: \begin{enumerate*}
    \item \emph{classifier and ensemble selection},
    \item \emph{functional} and
    \item \emph{non\-/functional ML adaptation}.
\end{enumerate*} At the end of this section, we also present a detailed comparison of our approach with the related work in terms of their category, objective, type of objective (functional/non functional), and applicability to ML models and properties.

\vspace{.5em}

\begin{table*}[!t]
    \caption{Comparison with related work. \label{tbl:related}}
    \begin{adjustbox}{max totalsize={.99\textwidth}{\textheight},center}

\footnotesize
\begin{tabular}{r | c | p{\widthof{\objSelection}} | c | c c}
    \toprule
    \multirow{2}{*}{\textbf{Ref.}} & \multirow{2}{*}{\textbf{Cat.}} & \multirow{2}{*}{\textbf{Objective}} & \multirow{2}{*}{\textbf{Objective Type}} & \multicolumn{2}{c}{\textbf{Applicability}}\\& & & & \textbf{ML Model} & \textbf{Property} \\
    \midrule

Cruz et al.~\cite{CRUZ20151925} 
    & \categorySelection
    & \objSelection
    & \objectiveTypeFunctional & \ok & \notok\\

Mousavi et al.~\cite{mousavi2018omni}
    & \categorySelection
    & \objSelection
    & \objectiveTypeFunctional & \ok & \notok\\

Roy et al.~\cite{ROY2018179}
    & \categorySelection
    & \objSelection
    & \objectiveTypeFunctional & \ok & \notok\\

P\'{e}rez-G\'{a}llego et al.~\cite{PEREZGALLEGO20191}
    & \categorySelection
    & \objSelection
    & \objectiveTypeFunctional & \ok & \notok\\

Zhang et al.~\cite{zhang4608310sv}
    & \categorySelection
    & \objSelection
    & \objectiveTypeFunctional & \ok & \notok\\

Zhu et al.~\cite{Zhu2023}
    & \categorySelection
    & \objSelection
    & \objectiveTypeFunctional & \ok & \notok\\

    \midrule

Almeida et al.~\cite{ALMEIDA201867}
    & \categoryAdaptationFunctional
    & \objSelection\ in the presence of drifts
    & \objectiveTypeFunctional & \ok & \notok\\

Tahmasbi et al.~\cite{pmlr-v139-tahmasbi21a} 
    & \categoryAdaptationFunctional
    & Select the best classifier in the presence of drifts
    & \objectiveTypeFunctional & \ok & \notok\\    

    \midrule

Iosifidis et al.~\cite{10.1007/978-3-030-27615-720} 
    & \categoryAdaptationNonFunctional
    & Fair predictions over time & \objectiveTypeNonFunctional & \ok & \soSo\\

Zhang et al.~\cite{10.5555/3367032.3367242, 9288346}
    & \categoryAdaptationNonFunctional
    & Balance quality and fairness over time 
    & \objectiveTypeFunctional, \objectiveTypeNonFunctional & \notok & \soSo\\

Iosifidis et al.~\cite{iosifidis2021online}
    & \categoryAdaptationNonFunctional
    & Fair predictions over time in online learning
    & \objectiveTypeNonFunctional & \soSo & \soSo\\

Zhang et al.~\cite{10.1007/978-3-030-75765-620}
    & \categoryAdaptationNonFunctional
    & Balance quality and fairness over time 
    & \objectiveTypeFunctional, \objectiveTypeNonFunctional & \notok & \soSo\\

Badar et al.~\cite{Badar2023}
    & \categoryAdaptationNonFunctional
    & Train a fair, Federated Learning model
    & \objectiveTypeNonFunctional & \notok & \soSo\\

    \midrule

    This & \NA & Stable application behavior over time wrt an arbitrary non\-/func. property
    & \objectiveTypeNonFunctional & \ok & \ok \\
    \bottomrule
\end{tabular}

     \end{adjustbox}
\end{table*}

\noindent \textbf{Classifier and ensemble selection} refers to the techniques that select the most suitable (set of) classifier among a set of candidates. It is referred to as \emph{classifier selection} when one classifier is selected, \emph{ensemble selection}, otherwise~\cite{CRUZ2018195}.
It can be performed at training time (\emph{static}), or for each (subset of) data point at inference time (\emph{dynamic}). The latter, often combined with static selection, typically shows the best performance~\cite{ROY2018179}. Selection maximizes functional metrics, often accuracy.
Meta\-/learning is frequently used, as surveyed by Khan et al.~\cite{8951014}.
For instance, Cruz et al.~\cite{CRUZ20151925} proposed a dynamic ensemble selection that considers different spatial\-/based criteria using a meta\-/classifier.
Zhu et al.~\cite{Zhu2023} defined a dynamic ensemble selection based on the generation of diversified classifiers. Selection is based on spatial information (i.e., the most \emph{competent} classifiers for a region). Classifiers predictions are combined using weighted majority voting, weights depend on the classifiers competency for a data point.
Zhang et al.~\cite{zhang4608310sv} defined a dynamic ensemble selection whose selection criterion considers the \emph{classifiers synergy}. It evaluates the \emph{contribution} of each classifier to the ensemble, in terms of the accuracy retrieved with and without the classifier. For each data point, it selects the classifiers with a positive contribution, and uses such contribution as weight in predictions aggregation.
Other approaches focused on imbalanced learning. Roy et al.~\cite{ROY2018179} showed that specific preprocessing (e.g., oversampling of the underrepresented class) and dynamic, spatial\-/based selection outperform static selection in this scenario. Mousavi et al.~\cite{mousavi2018omni} also used oversampling. Static selection then defines the ensemble and its combiner (e.g., majority voting). Dynamic selection finally retrieves a subset of the ensemble for each data point.
P\'{e}rez-G\'{a}llego et al.~\cite{PEREZGALLEGO20191} focused on \emph{quantification tasks} with drifts between classes. The proposed dynamic ensemble selection uses a specifically designed criterion, selecting the classifiers whose training distribution is the most similar to the input data points. 
Our approach implements a dynamic classifier selection, which departs from existing solutions implementing a (dynamic) selection of a (set of) classifier for each data point to maximize accuracy at inference time. Our goal is rather the \emph{run\-/time selection and substitution of the ML model to the aim of guaranteeing a stable behavior of the application with respect to a specific (set of) non\-/functional property}.

\vspace{.5em}

\noindent \textbf{Functional adaptation} refers to the techniques that adapt a ML model (and application) according to changing conditions, notably a drift, to keep quality metrics high.
According to the survey by Lu et al.~\cite{8496795}, the possible actions upon a detected drift are: training and using a new ML model, using ensemble purposefully trained for drift, and adapting an existing ML model when the drift is localized to a region.
The issue of drift has been approached using dynamic classifier selection. For instance, Almeida et al.~\cite{ALMEIDA201867} designed a drift detector whose selection criterion considers both spatial and concept\-/based information. It relies on a set of diverse classifiers that is dynamically updated, removing unnecessary classifiers and training new ones as new concepts emerge.
Tahmasbi et al.~\cite{pmlr-v139-tahmasbi21a} designed a novel adaptive ML model. It uses one classifier at time, and, upon drift detection, selects the subsequent classifier with the highest quality in the last evaluation window. 
Our approach implements an adaptation process, which departs from existing solutions based on the online re\-/training of individual ML models according to drift or the selection of ML models that maximize quality under drift.  Our goal is rather the \emph{adaptation of the overall ML-based application according to a (arbitrary) non\-/functional property of interest}.

\vspace{.5em}

\noindent \textbf{Non-functional adaptation} refers to the techniques that adapt a ML model (and application) according to a non\-/functional property. Fairness is the most studied property in literature in both static and dynamic settings; we focus on the latter due to its connection with the work in this paper.
For instance, Iosifidis et al.~\cite{10.1007/978-3-030-27615-720} designed an approach that tackles fairness and concept drift. It uses two pre\-/processing techniques modifying data, which are then taken as input by classifiers that can natively adapt to concept drifts (e.g., \emph{Hoefdding trees}). A similar solution is proposed by Badar et al.~\cite{Badar2023} in federated learning. It first detects drift, and then evaluates if fairness is no longer supported. It then performs oversampling as countermeasure.
Zhang et al.~\cite{10.5555/3367032.3367242, 9288346} introduced a training algorithm based on \emph{Hoefdding trees}, whose splitting criterion considers fairness and accuracy. Such idea has also been applied to random forest models~\cite{10.1007/978-3-030-75765-620}.
Iosifidis et al.~\cite{iosifidis2021online} designed an online learning algorithm that detects class imbalance and lack of fairness, and adjusts the ML model accordingly. It fixes weights during boosting (for imbalance) and the learned decision boundary (for fairness).
Our approach implements an adaptation process, which departs from existing re\-/training solutions  using a custom algorithm focused on a specific property (fairness).  Our goal is rather the \emph{adaptation of the overall application behavior according to any non\-/functional properties and ML algorithms}.

\vspace{.5em}

Table~\ref{tbl:related} shows how our approach compares with the related work in terms of \emph{\relatedWorkHeaderCategoryLong} (denoted as \relatedWorkHeaderCategoryShort), \emph{\relatedWorkHeaderObjLong}, \emph{\relatedWorkHeaderObjTypeLong}, and \emph{\relatedWorkHeaderGenericityLong}.
\emph{\relatedWorkHeaderCategoryLong} can be \begin{enumerate*}
    \item classifier and ensemble selection (denoted as \categorySelection),
    \item functional adaptation (denoted as \categoryAdaptationFunctional), and
    \item non\-/functional adaptation (denoted as \categoryAdaptationNonFunctional).
\end{enumerate*}
\emph{\relatedWorkHeaderObjTypeLong} can be \begin{enumerate*}
    \item functional (denoted as \objectiveTypeFunctional), and \item non\-/functional (denoted as \objectiveTypeNonFunctional). \end{enumerate*}
\emph{\relatedWorkHeaderGenericityLong} is expressed in terms of \begin{enumerate*}
    \item \emph{\relatedWorkHeaderGenericitySubModel} (\ok\ if applicable to any ML algorithm, \soSo\ if applicable to a class of ML algorithms, \notok\ if applicable to a specific ML algorithm only);
    \item \emph{\relatedWorkHeaderGenericitySubProperty} (\ok\ if applicable to any (non\-/)functional property, \soSo\ if applicable to a class of (non\-/)functional properties, \notok\ if applicable to a specific (non\-/)functional property).
\end{enumerate*}
Table~\ref{tbl:related} shows that 
our approach (last row in Table~\ref{tbl:related}) is the only \emph{architectural and methodological solution that supports stable non\-/functional behavior of ML\-/based applications}. It builds on a smart and dynamic multi\-/model substitution departing from expensive re\-/training approaches and inference-time classifier selection for individual data points. \section{Conclusions}\label{sec:conclusion}
We presented a multi\-/model approach for the continuous management of ML-based application non-functional behavior. 
Our approach guarantees a stable application behavior  at run time, over time and across model changes, where multiple ML models with similar non-functional properties are available and one model is selected at time according to such properties and the application context.  Our approach manages (dynamic and unpredictable) contextual changes in modern ML deployments, supporting early model substitutions based on Dynamic MAB and assurance evaluation. 

\section*{Acknowledgments}
Research supported, in parts, by \begin{enumerate*}
	\item project BA-PHERD, funded by the European Union -- NextGenerationEU, under the National Recovery and Resilience Plan (NRRP) Mission 4 Component 2 Investment Line 1.1: ``Fondo Bando PRIN 2022'' (CUP G53D23002910006);
    \item project MUSA -- Multilayered Urban Sustainability Action -- project, funded by the European Union - NextGenerationEU, under the National Recovery and Resilience Plan (NRRP) Mission 4 Component 2 Investment Line 1.5: Strengthening of research structures and creation of R\&D ``innovation ecosystems'', set up of ``territorial leaders in R\&D'' (CUP  G43C22001370007, Code ECS00000037);
    \item project SERICS (PE00000014) under the NRRP MUR program funded by the EU -- NextGenerationEU;
    \item 1H-HUB and SOV-EDGE-HUB funded by Universit\`a degli Studi di Milano -- PSR 2021/2022 -- GSA -- Linea 6; and
	\item program ``Piano di Sostegno alla Ricerca'' funded by Universit\`a degli Studi di Milano.
\end{enumerate*}
Views and opinions expressed are however those of the authors only and do not necessarily reflect those of the European Union or the Italian MUR. Neither the European Union nor the Italian MUR can be held responsible for them.

\bibliographystyle{IEEEtran}
\bibliography{biblio}

\begin{IEEEbiography}[{\includegraphics[width=1in,height=1.25in,clip,keepaspectratio]{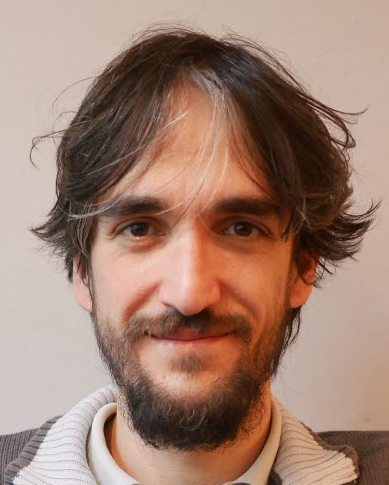}}]{Marco Anisetti}
	is Full Professor at the Department of Computer Science, Universit\`a degli Studi di Milano. His research interests are in the area of computational intelligence and its application to the design and evaluation of complex systems. He has been investigating innovative solutions in the area of assurance evaluation of cloud security and AI. In this area he defined a new scheme for continuous and incremental cloud security certification, based on distributed assurance evaluation architecture.
\end{IEEEbiography}

\vspace{-3em}

\begin{IEEEbiography}[{\includegraphics[width=1in,height=1.25in,clip,keepaspectratio]{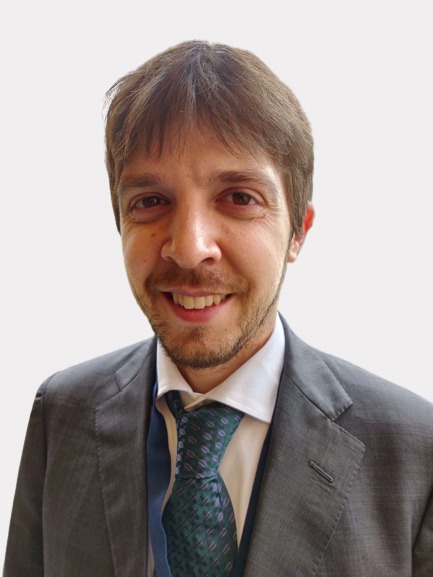}}]{Claudio A. Ardagna}
	is Full Professor at the Department of Computer Science, Universit\`a degli Studi di Milano, the Director of the CINI National Lab on Data Science, and co-founder of Moon Cloud srl. His research interests are in the area of edge-cloud and AI security and assurance, and data science. He has published more than 170 articles and books. He has been visiting professor at Universit\'e Jean Moulin Lyon 3 and visiting researcher at BUPT, Khalifa University, GMU.
\end{IEEEbiography}

\vspace{-3em}

\begin{IEEEbiography}[{\includegraphics[width=1in,height=1.25in,clip,keepaspectratio]{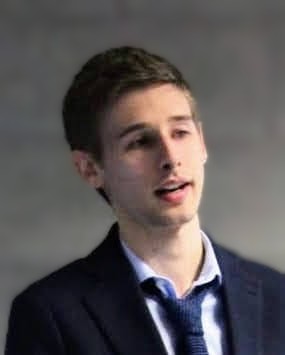}}]{Nicola Bena} 
	is a postdoc at the Department of Computer Science, Universit\`a degli Studi di Milano. His research interests are in the area of security of modern distributed systems with particular reference to certification, assurance, and risk management techniques. He has been visiting scholar at Khalifa University and at INSA Lyon.
\end{IEEEbiography}

\vspace{-3em}

\begin{IEEEbiography}[{\includegraphics[width=1in,height=1.25in,clip,keepaspectratio]{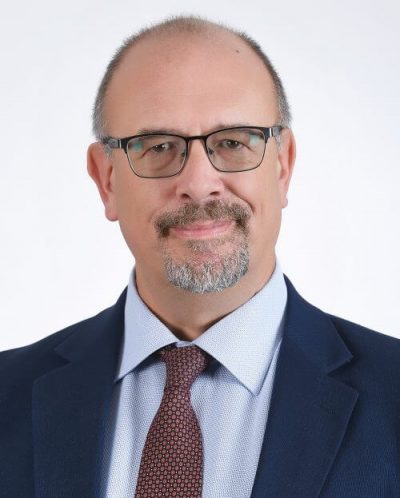}}]{Ernesto Damiani}
is Full Professor at the Department of Computer Science, Universit\`a degli Studi di Milano, where he leads the Secure
	Service-oriented Architectures Research (SESAR) Laboratory. 
	He is also the Founding Director of the Center for Cyber-Physical Systems, Khalifa University, UAE. He received an Honorary Doctorate from INSA Lyon for his contributions to research and teaching on big data analytics.  His research interests include cybersecurity, big data, artificial intelligence, and cloud/edge processing, and he has published over 680 peer-reviewed articles and books. He is a Distinguished Scientist of ACM and was a recipient of the 2017 Stephen Yau Award.
\end{IEEEbiography}

\vspace{-3em}

\begin{IEEEbiography}[{\includegraphics[width=1in,height=1.25in,clip,keepaspectratio]{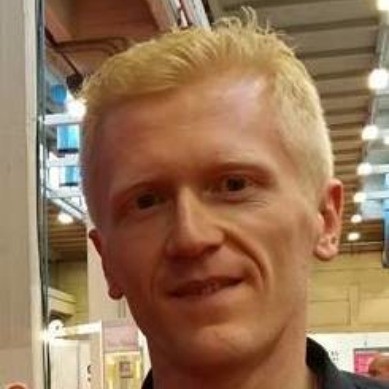}}]{Paolo G. Panero} is a master student at the Department of Computer Science, Universit\`a degli Studi di Milano. He is IT officer in a in-house public administration company where he deals with innovation and IT services. His research interests are in the area of Machine Learning with a focus on models evaluation.
\end{IEEEbiography}

\end{document}